\documentclass{article} 
\pdfoutput=1
\usepackage{nips15submit_e,times}
\usepackage{url}

\usepackage{color}

\usepackage{amsmath}
\usepackage{amssymb}
\usepackage{booktabs}
\usepackage{pifont}
\usepackage[utf8]{inputenc}
\usepackage{float}
\usepackage{flushend}

\usepackage{multirow}
\usepackage{multicol}
\usepackage{tabularx}
\usepackage{cite}
\usepackage{enumitem}

\usepackage[pdftex]{graphicx}
\usepackage{epstopdf}
   \graphicspath{{../images/}}
   \DeclareGraphicsExtensions{.pdf,.jpeg,.png,.eps}

\usepackage[]{algorithm2e,setspace}
\makeatletter
\renewcommand{\@algocf@capt@plain}{above}
\makeatother
\newcommand*\rot{\rotatebox{90}}
\newcommand*\OK{\ding{51}}

\newif\ifCommentsAuthors

\CommentsAuthorsfalse

\ifCommentsAuthors
  
    \definecolor{myred}{rgb}{.8,.0,.0}
    
    \definecolor{myblue}{rgb}{0,0,.8}
    \newcommand{\comment}[1]{\textcolor{myblue}{#1}}
    \definecolor{mcolor}{rgb}{0,0.5,0.1}
    \newcommand{\commentm}[1]{\textcolor{mcolor}{#1}}

\else

    \definecolor{myred}{rgb}{.8,.0,.0}
    
    \newcommand{\commentm}[1]{}
    \newcommand{\comment}[1]{}
    
\fi

\title{Multiple Instance Learning: A Survey of Problem Characteristics and Applications}

\author{
Marc-André Carbonneau\thanks{Laboratoire d'imagerie, de vision et d'intelligence artificielle, École de technologie supérieure, Montreal, Canada}\\
{marcandre.carbonneau@gmail.com}
\and
\textbf{Veronika~Cheplygina}\thanks{
Biomedical Imaging Group Rotterdam, Erasmus Medical Center, Rotterdam, The Netherlands and Pattern Recognition Laboratory, Delft University of Technology, Delft, The Netherlands}\\
{v.cheplygina@tudelft.nl}
\AND
\textbf{Eric~Granger}\footnotemark[1]\\
{eric.granger@etsmtl.ca}
\and
\textbf{Ghyslain~Gagnon}\thanks{Laboratoire de communications et d'intégration de la microélectronique, École de technologie supérieure, Montreal, Canada}\\
{ghyslain.gagnon@etsmtl.ca}
}

\nipsfinalcopy 

\begin{document}

\maketitle

\begin{abstract}
Multiple instance learning (MIL) is a form of weakly supervised learning where training instances are arranged in sets, called bags, and a label is provided for the entire bag. This formulation is gaining interest because it naturally fits various problems and allows to leverage weakly labeled data. Consequently, it has been used in diverse application fields such as computer vision and document classification. However, learning from bags raises important challenges that are unique to MIL. This paper provides a comprehensive survey of the characteristics which define and differentiate the types of MIL problems. Until now, these problem characteristics have not been formally identified and described. As a result, the variations in performance of MIL algorithms from one data set to another are difficult to explain. In this paper, MIL problem characteristics are grouped into four broad categories: the composition of the bags, the types of data distribution, the ambiguity of instance labels, and the task to be performed. Methods specialized to address each category are reviewed. Then, the extent to which these characteristics manifest themselves in key MIL application areas are described. Finally, experiments are conducted to compare the performance of 16 state-of-the-art MIL methods on selected problem characteristics.  This paper provides insight on how the problem characteristics affect MIL algorithms, recommendations for future benchmarking and promising avenues for research. 
\end{abstract}

\section{Introduction}

Multiple instance learning (MIL) deals with training data arranged in sets, called bags. Supervision is provided only for entire sets, and the individual label of the instances contained in the bags are not provided. This problem formulation has attracted much attention from the research community, especially in the recent years, where the amount of data needed to address large problems has increased exponentially. Large quantities of data necessitate a growing labeling effort. 

Weakly supervised methods, such as MIL, can alleviate this burden since weak supervision is generally obtained more efficiently. For example, object detectors can be trained with images collected from the web using their associated tags as weak supervision, instead of locally-annotated data sets~\cite{Hoffman2015,Wu2015deep}. Computer-aided diagnosis algorithms can be trained with medical images for which only patient diagnoses are available instead of costly local annotations provided by an expert. Moreover, there are several types of problems that can naturally be formulated as MIL problems. For example, in the drug activity prediction problem \cite{Dietterich1997}, the objective is to predict if a molecule induces a given effect. A molecule can take many conformations which can either produce, or not, a desired effect. Observing the effect of individual conformations is unfeasible. Therefore, molecules must be observed as a group of conformations, hence use the MIL formulation. Because of these attractive properties, MIL has been increasingly used in many other application fields over the last 20 years, such as image and video classification \cite{Chen2006,Rahmani2006,Andrews02,Zhang2002CBIR,Phan2015,Cinbis2016}, document classification \cite{Zhou2009migraph,Bunescu2007relation} and sound classification \cite{Briggs2012}.

Several comparative studies and meta-analyses have been published to better understand MIL \cite{Zhou2004survey,Babenko2008,Amores2013,Doran2014,Foulds2010,Ray2005,Cheplygina2015Group,Vanwinckelen2015,Alpaydin2015,Cheplygina2015stab,cheplygina2015characterizing}. All these papers observe that the performance of MIL algorithms depends on the characteristics of the problem. While some of these characteristics have been partially analyzed in the literature~\cite{Li2010convex,Bunescu2007relation,Zhou2009migraph,han2010}, a formal definition of key MIL problem characteristics has yet to be described. 

A limited understanding of such fundamental problem characteristics affects the advancement of MIL research in many ways. Experimental results can be difficult to interpret, proposed algorithms are evaluated on inappropriate benchmark data sets, and results on synthetic data often do not generalize to real-world data. Moreover, characteristics associated with MIL problems have been addressed under different names. For example, the scenario where the number of positive instances in a bag is low was referred to as either sparse bags~\cite{Yan2016,Bunescu2007} or low witness rate~\cite{Li2013,Li2010convex}. It is thus important for future research to formally identify and analyze what defines and differentiates MIL problems. 

\commentm{What is in the survey?}
This paper provides a comprehensive survey of the characteristics inherent to MIL problems, and investigates their impact on the performance of MIL algorithms. These problem characteristics are all related to unique features of MIL: the ambiguity of instance labels and the grouping of data in bags. We propose to organize problem characteristics in four broad categories: \textit{Prediction level}, \textit{Bag composition}, \textit{Label ambiguity} and \textit{Data distribution}.

Each characteristic raises different challenges. When instances are grouped in bags, predictions can be performed at two levels: bags-level or instance-level~\cite{Cheplygina2015Group}. Algorithms are often better suited for only one of these two types of task~\cite{Vanwinckelen2015,Alpaydin2015}. Bag composition, such as the proportion of instances from each class and the relation between instances, also affects the performance of MIL methods. The source of ambiguity on instance labels is another important factor to consider. This ambiguity can be related to label noise as well as to instances not belonging to clearly defined classes \cite{Foulds2010}. Finally, the shape of positive and negative distributions affects MIL algorithms depending on their assumptions about the data. 

As additional contributions, this paper reviews state-of-the-art methods which can address challenges of each problem characteristic. It also examines several applications of MIL, and in each case, identifies their main characteristics and challenges. For example, in computer vision, instances can be spatially related, but this relationship does not exist in most bioinformatics applications. Finally, experiments show the effects of selected problem characteristics -- the instance classification task, witness rate, and negative class modeling -- with 16 representative MIL algorithms. This is the first time that algorithms are compared on the bag and instance classification tasks in the light of these specific challenges. Our findings indicate that these problem characteristics have a considerable impact on the performance of all MIL methods, and that each method is affected differently. Therefore, problem characterization cannot be ignored when proposing new MIL methods and conducting comparative experiments. Finally, this paper provides novel insights and direction to orient future research in this field from the problem characteristics point-of-view.

The rest of this paper is organized as follows. The next section describes MIL assumptions and the different learning tasks that can be performed using the MIL framework. Section \ref{Section:Surveys} reviews previous surveys and general MIL studies. Section \ref{Section:keyProblems} and \ref{Section:Applications} identify and analyze the key problem characteristics and applications, respectively. Experiments are presented in Section \ref{Section:Experiments}, followed by a discussion in Section \ref{Section:Discussion}.

\section{Multiple Instance Learning}
\label{Section:MIL}

\subsection{Assumptions}
In this paper, two broad assumptions are considered: the standard and the collective assumption. For a more detailed review on the subject, the reader is referred to \cite{Foulds2010}.

The \textit{standard MIL assumption} states that all negative bags contain only negative instances, and that positive bags contain at least one positive instance. These positive instances are named witnesses in many papers and this designation is used in this survey. Let $Y$ be the label of a bag $X$, defined as a set of feature vectors $X = \left \{ \textbf{x}_{1}, \textbf{x}_{2}, ..., \textbf{x}_{N}  \right \}$. Each instance (i.e. feature vector) $\textbf{x}_i$ corresponds to a label $y_i$. The label of the bag is given by:  
\begin{equation}
Y = \left\{\begin{matrix}
+1 & \text{if} & \exists y_i :\; y_{i}=+1; \\ 
-1 & \text{if} & \forall y_i :\; y_{i}=-1.
\end{matrix}\right. 
\end{equation} This is the working assumption of many of the early methods \cite{Dietterich1997,Andrews02,Maron1998}, as well as recent ones \cite{Carbonneau2016,Xiao2016}. To correctly classify bags under the standard assumption, it is not necessary to identify all witnesses as long as at least one is found in each positive bag. This will be discussed in detail in Section \ref{Section:IvsB}.    

The standard MIL assumption can be relaxed to address problems where positive bags cannot be identified by a single instance, but by the interaction or the accumulation of several instances. A simple representative example given by Foulds and Frank \cite{Foulds2010} is the classification of desert, sea and beach images. Images of deserts will contain sand segments, while images of the sea contain water segments. However, images of beaches must contain both types of segments. To correctly classify beach images, the model must verify the presence of both types of witnesses, and thus, methods working under the standard MIL assumption would fail in this case. In some problems, several positive instances are necessary to assign a positive label to a bag. For example, in traffic jam detection from images of a road, a car would be a positive instance. However, it takes many cars to create a traffic jam. 
In this survey, the \textit{collective assumption} designates all assumptions in which more than one instance defines bag labels. 

\subsection{Tasks}
\label{Section:Tasks}

\textbf{Classification:}
Classification can be performed at two levels: bag and instance. Bag classification is the most common task for MIL algorithms. It consists in assigning a class label to a set of instances. The individual instance labels are not necessarily important depending on the type of algorithm and assumption. Instance classification is different from bag classification because while training is performed using data arranged in sets, the objective is to classify instance individually. As pointed out in \cite{Carbonneau2016IPTA}, the loss functions for the two tasks are different (see Section \ref{Section:IvsB}). When the goal is bag classification, misclassifying an instance does not necessarily affect the loss at bag-level. For example, in a positive bag, few true negative instances can be erroneously classified as positive and the bag label will remain unchanged. Thus, the structure of the problem, such as the number of instances in bags, plays an important role in the loss function \cite{Vanwinckelen2015}. As a result, the performance of an algorithm for bag classification is not representative of the performance obtained for instance classification. Moreover, many methods proposed for bag classification (e.g. \cite{Zhang2001,Zhou2007CCE}) do not reason in instance space, and thus, often cannot perform instance classification. 

MIL classification is not limited to assigning a single label to instances or bags. Assigning multiple labels to bags is particularly relevant considering that they can contain instances representing different concepts. This idea has been the object of several publications \cite{Herrera2016MLMIL}. Multi-label classification is subject to the same problem characteristics as single label classification, thus no distinction will be made between the two in the rest of this paper.



\textbf{Regression:}
MIL regression task consists in assigning a real value to a bag (or an instance) instead of a class label. The problem has been approached in different ways. Some methods assign the bag label based on a single instance. This instance may be the closest to a target concept \cite{Dooly2003}, or the best fit in a regression model \cite{Ray2001regression}. Other methods work under the collective assumption and use the average or a weighted combination of the instances to represent bags as a single feature vector \cite{Wang2008aerosol,Wagstaff2007,Pappas2014}. Alternatively, on can simply replace a bag-level classifier by a regressor \cite{Manzalawy2011}. 

\textbf{Ranking:}
Some methods have been proposed to rank bags or instances instead of assigning a class label or a score. The problem differs from regression because the goal is not to obtain an exact real valued label, but to compare the magnitude of scores to perform sorting. Ranking can be performed at the bag level \cite{Bergeron2012} or at the instance level \cite{Hu2008rank}.

\textbf{Clustering:}
This task consists in finding clusters or a structure among a set of unlabeled bags. The literature on the subject is limited. In some cases, clustering is performed in bag space using standard algorithms and set-based distance measures (e.g. \textit{k}-Medoids and the Hausdorff distance \cite{Zhang2009Clust}). Alternatively, clustering can be performed at the instance level. For example, in \cite{Zhang2011M3IC}, the algorithm identifies the most relevant instance of each bag, and performs maximum margin clustering on these instances. 
 


Most of the discussion in the remainder of the paper will be articulated around classification, as it is the most studied task. However, challenges and conclusions related to problem characteristics are also applicable to the other tasks.

\section{Studies on MIL}
\label{Section:Surveys}
Because many problems can be formulated as MIL, there is a plethora of MIL algorithms in the literature. However, there is only a handful of general MIL studies and surveys. This section summarizes and interprets the broad conclusions from these general MIL papers.

\commentm{General Surveys}
The first survey on MIL is a technical report written in 2004 \cite{Zhou2004survey}. It describes several MIL algorithms, some applications and discusses learnability under the MIL framework. In 2008, Babenko published a report \cite{Babenko2008} containing an updated survey of the main families of MIL methods, and distinguished two types of ambiguity in MIL problems. The first type is polymorphism ambiguity, in which each instance is a distinct entity or a distinct version of an entity (e.g. conformations of a molecule). The second is part-whole ambiguity in which all instances are parts of the same object (e.g. segments of an image). In a more recent survey \cite{Amores2013}, Amores proposed a taxonomy in which MIL methods are divided in three broad categories following the representation space. Methods operating in the instance space are grouped together, and the methods operating in bag space are divided in two categories based on whether a bag embedding is performed or not. Several experiments were performed to compare bag classification accuracy in four application fields. Bag-level methods performed better in terms of bag classification accuracy, however, performance depends on the data and the distance function or the embedding method. Finally, very recently, a book on MIL has been published \cite{Herrera2016MILBook}. It discusses most of the tasks of Section \ref{Section:Tasks} along with associated methods, as well as data reduction and imbalanced data.   

\commentm{Studies on general properties of MIL}
Some papers study specific topics of MIL. For instance, Foulds and Frank reviewed the assumptions \cite{Foulds2010} made by MIL algorithms. They stated that these assumptions influence how algorithms perform on different types of data sets. They found that algorithms working under the collective assumption also perform well with data sets corresponding to the standard MIL assumption, such as the Musk data set \cite{Dietterich1997}. Sabato and Tishby \cite{Sabato2012} analyzed the of sample complexity in MIL, and they found that the statistical performance of MIL is only mildly dependent on the number of instances per bag. In \cite{cheplygina2015characterizing} the similarities between MIL benchmark data sets were studied. The data sets were represented in two ways: by meta-features describing numbers of bags, instances and so forth, and by features based on performances of MIL algorithms. Both representations were embedded in a 2-D space and found to be dissimilar to each other. In other words, data sets often considered similar due to the application or size of data did not behave similarly, which suggest that some unobserved properties influence MIL algorithm performances.

\commentm{Comparison with other types of learning}
Some papers compare MIL to other learning settings to better understand when to use MIL. Ray and Craven \cite{Ray2005} compared the performance of MIL methods against supervised methods on MIL problems. They found that in many cases, supervised methods yield the most competitive results. They also noted that, while some methods systematically dominate others, the performance of the algorithms was application-dependent. In \cite{Cheplygina2015Group}, the relationship between MIL and settings such as group-based classification and set classification is explored. They state that MIL is applicable in two scenarios: the classification of bags and the classification of instances. Recently, these differences were rigorously investigated \cite{Vanwinckelen2015}. It was shown analytically and experimentally that the correlation between classification performance at bag and instance level is relatively weak. Experiments showed that depending on the data set, the best algorithm for bag classification provides average, or even the worst performance for instance classification. They too observed that different MIL algorithms perform differently given the nature of the data.

\commentm{Instance-space methods}
The classification of instances can be a task in itself, but can also be an intermediate step toward bag classification for instance space methods \cite{Amores2013}. Alpaydin et al. \cite{Alpaydin2015} compared instance-space and bag-space classifiers on synthetic and real-world data. They concluded that for datasets with few bags, it is preferable to use an instance-level classifier. They also state, as in \cite{Amores2013}, that if the instances provide partial information about the bag labels, it is preferable to use bag-level representation. In \cite{Cheplygina2015stab}, Cheplygina et al. explored the stability of the instance labels assigned by MIL algorithms. They found that algorithms yielding best bag classification performance were not the algorithms providing the most consistent instance labels. Carbonneau et al.~\cite{Carbonneau2016ICPR} studied the ability to identify witnesses (positive instances) of several MIL methods. They found that depending on the nature of the data, some algorithms perform well while others would have difficulty learning.

\commentm{Specific methods and applications}
Finally, some papers focus on specific classes of algorithms and applications. Doran and Ray \cite{Doran2014} analyzed and compared several SVM-based MIL methods. They found that some methods perform better for instance classification than for bag classification, or vice-versa, depending on the method properties. Wei and Zhou \cite{Wei2016} compared methods for generating bags of instances from images. They found that sampling instances densely leads to a higher accuracy than sampling instances at interest points or after segmentation. This agrees with other bag-of-words (BoW) empirical comparisons \cite{Nowak2006,Wang2009}. They also found that methods using the collective assumption performed better for image classification. Vankatesan et al. \cite{Venkatesan2015} showed that simple lazy-learning techniques could be applied to some MIL problems to obtain results comparable to state-of-the-art techniques. Kandemir and Hamprecht \cite{Kandemir2015} compared several MIL algorithms in two computer-aided diagnosis (CAD) applications. They found that modeling intra-bag similarities was a good strategy for bag classification in this context.


The main conclusions of these studies are summarized as follows:
\begin{itemize}
\item The performance of MIL algorithms depends on several properties of the data set \cite{Amores2013,Ray2005,Vanwinckelen2015,Alpaydin2015,Carbonneau2016ICPR,cheplygina2015characterizing}.
\item When it is necessary to model combinations of instances to infer bag labels, bag-level and embedding methods perform better \cite{Alpaydin2015,Amores2013,Wei2016}.
\item The best bag-level classifier is rarely the best instance-level classifier, and vice versa \cite{Vanwinckelen2015,Doran2014}.
\item When the number of bags is low, it is preferable to use an instance-based method \cite{Alpaydin2015}.
\item Some MIL problems can also be solved using standard supervised methods \cite{Ray2005}. 
\item Performance of MIL is only mildly dependent on the number of instances per bag \cite{Sabato2012}.
\item Similarity between the instances of a same bag affect classification performance \cite{Kandemir2015}. 
\end{itemize}

All of these conclusions are related to one or more characteristics that are unique to MIL problems. \textbf{Identifying these characteristics and gaining a better understanding of their impact on MIL algorithms is an important step towards the advancement of MIL research.}

\section{Characteristics of MIL Problems}
\label{Section:keyProblems}
\begin{figure}
\centering
\includegraphics[width=1.0\textwidth]{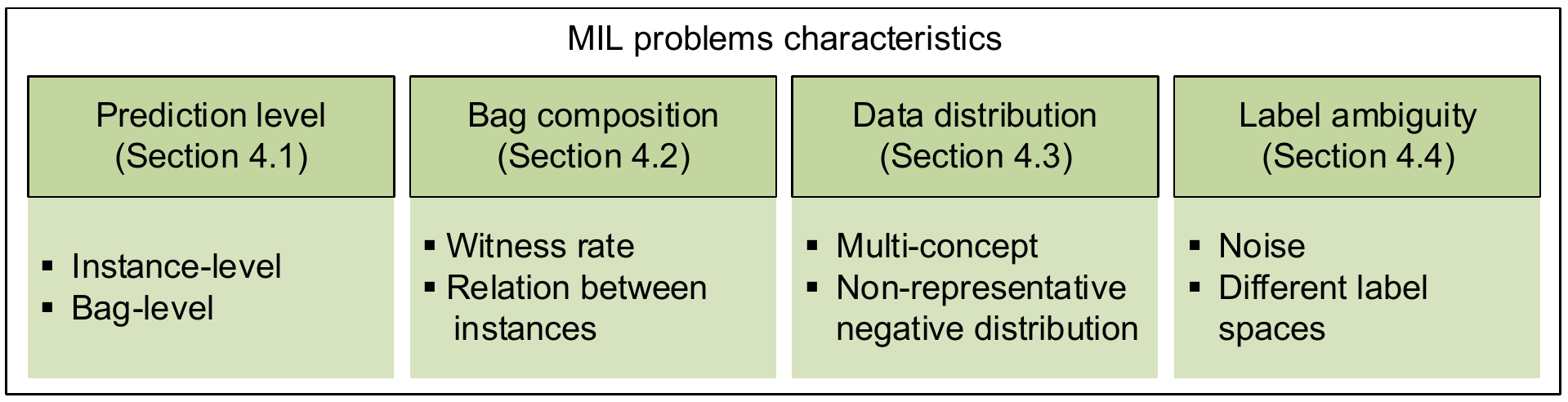}
\caption{Characteristics inherent to MIL problems.}
\label{Figure:Tree}    
\end{figure}

We identified four broad categories of key characteristics associated with MIL problems which directly impacts on the behavior of MIL algorithms: \textit{task}, \textit{bag composition}, \textit{data distributions} and \textit{label ambiguity} (as shown in Fig. \ref{Figure:Tree}). Each characteristic poses different challenges which must be addressed specifically.

In the remainder of this section, each of these characteristics will be discussed in more detail, along with representative specialized methods proposed in the literature to address them.

\subsection{Prediction: Instance-level vs. Bag-level}
\label{Section:IvsB}
In some applications, like object localization in images, the objective is not to classify bags, but to classify individual instances. While these two tasks appear similar, there are key differences, and thus, the bag classification performance of a method often is not representative of its instance classification performance \cite{Vanwinckelen2015,Doran2014}. It was shown in analytic and empirical investigations \cite{Vanwinckelen2015} that the relationship between the accuracy at the two levels depends of the number of instances in bags, the class imbalance and the accuracy of the instance classifier. This means that algorithms designed for bag classification are not optimal for instance classification. Most methods in the literature address the bag classification problem, and sometimes perform instance classification as a \textit{side feature} (e.g. MILES \cite{Chen2006}). One of the challenges for developing instance-level classification algorithm is the scarcity of benchmark data sets providing ground truth for instance labels.

\begin{figure}
\centering
\includegraphics[width=0.65\textwidth]{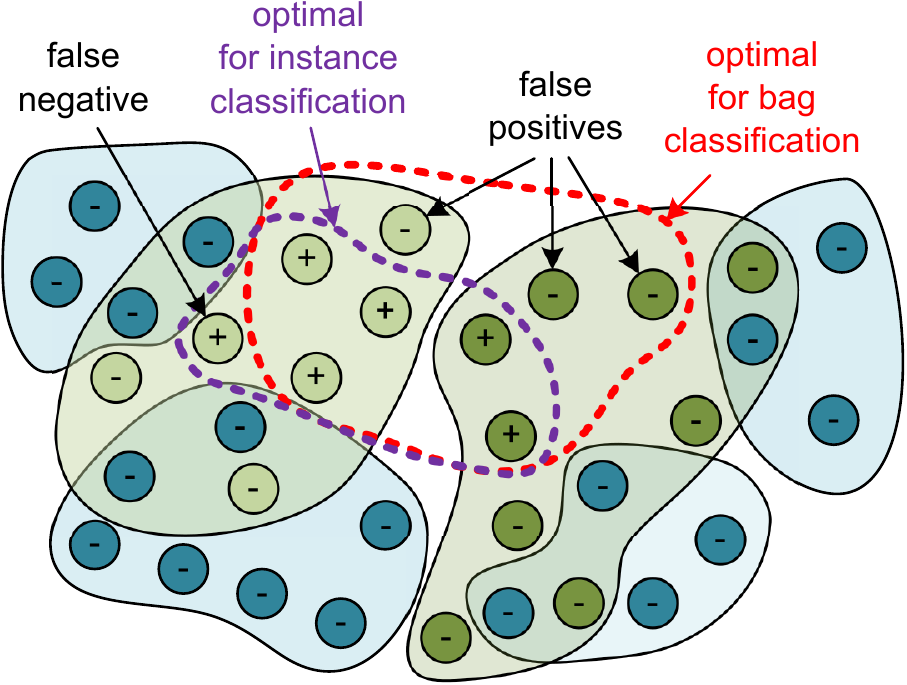}
\caption{Illustration of two decisions boundaries on a fictive problem. While only the purple boundary correctly classifies all instances, both them achieve perfect bag classification. This is because, in that case, false positive and false negative instances do not impact on bag labels.}
\label{Figure:TH}    
\end{figure}

The main difference between the two tasks is the misclassification cost of instances. Under the standard MIL assumption, as soon as a witness is identified in a bag, it is labeled as positive and all other instance labels can be ignored. In that case, false positives (FP) and false negatives (FN) have no impact on the bag classification accuracy, but still count as classification errors at the instance level. In addition, when considering negative bags, a single FP causes a bag to be misclassified. This means that if 1\% of the instances in each negative bag were misclassified, the accuracy on negative bags would be 0\%, although the accuracy on negative instances would be 99\%. This is illustrated in Fig. \ref{Figure:TH}. The green ensembles represent positive bags, while negative bags correspond to blue ensembles. The individual labels of the instances are identified on each instance. In this figure, both decision boundaries (dotted lines) are optimal for bag classification because they include at least one instance from all positive bags, while excluding all instances from negative bags. However, only one of the two boundaries achieves perfect instance classification (purple). This is why MIL algorithms using bag accuracy as an optimization criterion (e.g. APR \cite{Dietterich1997}, MI-SVM \cite{Andrews02}, MIL-Boost \cite{Babenko2008alignement}, EM-DD \cite{Zhang2001}, MILD \cite{Li2010MILD}) can learn a suboptimal decision boundary for instance classification. 

It has been proposed to consider negative and positive bags separately in the classifier loss function \cite{Jia2008}. The accuracy on positive bags is taken at bag level, but for negative bags, all instances are treated individually. This optimization criterion was proposed to adjust the decision threshold of bag classifiers for instance classification and improve their accuracy in \cite{Carbonneau2016IPTA}. In \cite{Yang2006asym}, a different weight is attributed to FP and FN during the optimization of an SVM. Some methods label all instances independently, like mi-SVM \cite{Andrews02} and MissSVM \cite{Zhou2007Semi}. These methods yield the best results in our experiments on instance-level classification (see Section \ref{Section:ExpIC}).

\subsection{Bag Composition}
\label{Section:BagComposition}

\subsubsection*{Witness Rate}
The witness rate (WR) is the proportion of positive instances in positive bags. When the WR is very high, positive bags contain only a few negative instances. In that case, the label of the instances can be assumed be the same as the label of their bag. The problem then reverts to a supervised problem with one-sided noise which can be solved in a regular supervised framework \cite{Blum1998}. However, in some applications, WR can be arbitrarily small and hinder the performance of many algorithms. For example, in methods like Diverse Density (DD) \cite{Maron1998}, Citation-kNN \cite{Zhang2001} and APR \cite{Dietterich1997} instances are considered to have the same label as their bag. When the WR is low, this is no longer reasonable and leads to lower performances. Methods which analyze instance distributions in bags \cite{Amores2010,Doran2014dist,Wei2014miFV} may also have problems dealing with low WR because distribution in positive and negative bags become similar. Also, some methods represent bags by the average of the instances they contain, like NSK-SVM \cite{Gartner2002}, or by considering their contribution to the bag label equally \cite{Xu2004LR}. With very low WRs, the few positive instances have a limited effect after the pooling process. Finally, in instance classification problems, lower WRs mean serious class imbalance problems, which leads to bad performance for many methods.

Several authors studied low WR problems in recent years. For example, sparse transductive MIL (stMIL) \cite{Bunescu2007} is an SVM formulation similar to NSK-SVM \cite{Gartner2002}. However, to better deal with low WR bags, the optimization constraints of the SVM are modified to be satisfied when at least one witness is found in positive bags. This method performs well at low WR but is less efficient when it is higher. Sparse balanced MIL (sbMIL) \cite{Bunescu2007} incorporates an estimation of the WR as a parameter in the optimization objective to solve this problem. WR estimation has also been successfully used in low WR problems by ALP-SVM \cite{Gehler2007}, SVR-SVM \cite{Li2010convex} and the $\gamma$-rule \cite{Li2013}. One drawback of using the WR as a parameter is that the WR is assumed to be constant across all bags. Other methods, like CR-MILBoost \cite{Ali2014} and RSIS \cite{Carbonneau2016}, estimate the probability that each instance is positive before training an ensemble of classifiers. During training, the classifiers give more importance to the instances that are more likely to be witnesses. In miGraph \cite{Zhou2009migraph}, similar instances in a bag are grouped in cliques. The importance of each instance is inversely proportional to the size of its clique. Assuming positive and negative instances belong to different cliques, the WR has little impact. In miDoc \cite{Yan2016}, a graph represents the entire MIL problem, where bags are compared based on the connecting edges. Experiments show that the method performs well on very low WR problems.

\subsubsection*{Relations Between Instances}

Most existing MIL methods assume, often not explicitly, that positive and negative instances are sampled independently from a positive and a negative distribution. However, this is rarely the case with real-world data. In many applications, the i.i.d. assumption is violated because structure or correlations exist between the instances and bags \cite{Zhou2009migraph,Zhang2011MILSD}. We make a distinction between three types of relation: intra-bag similarities, instance co-occurrences and structure.

\begin{figure}
\centering
\includegraphics[width=0.35\textwidth]{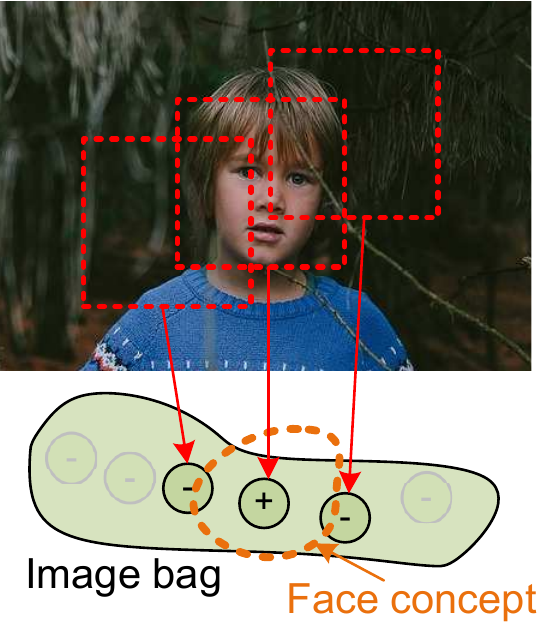}
\caption{Illustration of intra-bag similarity between instances: The patches are overlapping, and thus, share similarities with each other.}           
\label{Figure:kid}       
\end{figure}

\begin{figure}
\centering
\includegraphics[width=0.45\textwidth]{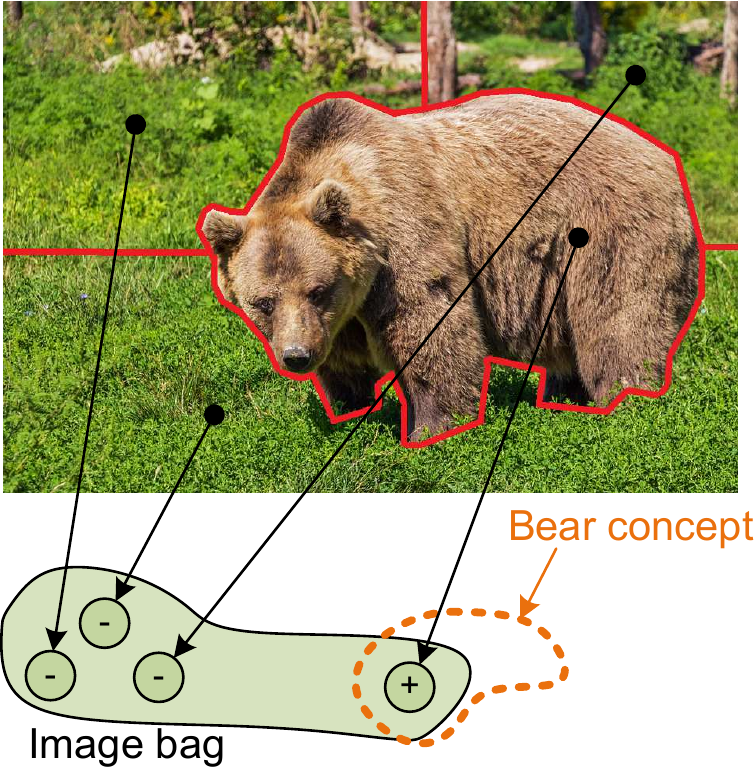}
\caption{Example of co-occurrence and similarity between instances: Three segments contain grass and forest and are therefore very similar. Moreover, since this is an image of a bear, the background is more likely to be nature than a nuclear central control room.}              
\label{Figure:bear}
\end{figure}

\textbf{Intra-Bag Similarities: }
In some problems, the instances belonging to the same bag share similarities, that  instances from other bags do not share. For instance, in the drug activity prediction problem \cite{Dietterich1997}, each bag contains many conformations of the same molecule. It is likely that instances of the same molecule are similar to some extent, while being different from other molecules \cite{Zhou2004survey}. One must thus ensure that the MIL algorithm learns to differentiate active from non-active conformations, instead of learning to classify molecules. In image-related applications, it is likely that all segments share some similarities related the capture condition (e.g. illumination, noise, etc.). Alternatively, similarities between instances of a same bag may be related to the instance generation process. For example, some methods use densely extracted patches which overlap (Figure \ref{Figure:kid}). Since they share a certain number of pixels, they are likely to be correlated.  Also, the background of a picture could be split in different segments which can be very similar (see Figure \ref{Figure:bear}). 

Intra-bag similarities raise some difficulties when learning. For instance, transductive algorithms (e.g. mi-SVM \cite{Andrews02}) might not be able to infer instance labels if the negative instances from positive and negative bags differ in nature \cite{Ray2005}.

Very few methods were proposed explicitly to address this problem. To deal with similar instances, miGraph \cite{Zhou2009migraph} builds a graph per bag and groups similar instances together to adjust their relative importance based on the group size. In CCE \cite{Zhou2007CCE}, a binary vector represents the bags by encoding the assignation of at least one instance to a cluster. Because features are binary, many instances can be assigned to the same cluster and the representation remains unaffected, which provides robustness to intra-bag similarity.

\textbf{Instance Co-occurrence: }
Instances co-occur in bags when they share a semantic relation. This type of correlation happens when the subject of a picture is more likely to be seen in some environment than in another, or when some objects are often found together (e.g. knife and fork). For example, the bear of Figure \ref{Figure:bear} is more likely to be found in nature than in a nightclub. Thus, the observation of nature segments might help to decide if the image contains a cocktail or a bear \cite{Kang2006}. In \cite{Cheplygina2015Diss}, it is shown that different birds are often heard in the same audio fragment, so a ``negative'' bird song could help to correctly classify the bird of interest. In these examples, co-occurrence represents an opportunity for better accuracy, however, in some cases it is a necessary condition for successful classification. Consider the example given by Foulds and Frank \cite{Foulds2010} where one must classify sea, desert and beach images. Both desert and beach images can contain sand instances, while water instances can be found in sea and beach images. However, both instances must co-occur in a beach image. Most methods working under the collective assumption \cite{Foulds2010} naturally leverage co-occurrence. Many of these methods, like BoW \cite{Amores2010,Csurka2004}, miFV \cite{Wei2014miFV}, FAMER \cite{Ping2011FAMER} or PPMM \cite{Wang2008PPMM} represent bags as instance distributions which indirectly account for co-occurrence. This has also been directly modeled in a tensor model \cite{Qi2007} and in a multi-label framework \cite{Zha2008}.

While useful to classify bags, in instance classification problems, the co-occurrence of instances may confuse the learner. If a given positive instance often co-occurs with a given negative instance, the algorithm is more likely to consider the negative instance as positive, which in this context would lead to a higher false positive rate (FPR).

\textbf{Instance and Bag Structure: }
In some problems, there exists an underlying structure between instances in bags or even between bags \cite{Zhang2011MILSD}. Structure is more complex than simple co-occurrence in the sense that instances follow a certain order, or are related in a meaningful way. Capturing this structure may lead to better classification performance \cite{Zhou2009migraph,laptev2008,Ryoo2009}. The structure may be spatial, temporal, relational or even causal. For example, when a bag represents a video sequence, all frames or patches are temporally and spatially ordered. For example, it is difficult to differentiate between a person taking or leaving a package without taking this temporal order into account. Alternatively, in web mining tasks \cite{Zhang2011MILSD} where websites are bags and pages linked by the websites are instances, there exists a semantic relation between two bags representing websites linked together.

Graph models were proposed to better capture the relations between the different entities in non-i.i.d. MIL problems to increase classification performance. Structure can be exploited at many levels: graphs can be used to model the relations between bags, instances or both \cite{Yan2016,Zhang2011MILSD}. Graphs enforce that related objects belong to the same class. Alternatively, in \cite{Mcgovern2003} bags are represented by a graph capturing diverse relationships between objects. The objects are shared across all bags and all possible sub-graphs of the bag graph correspond to instances.

Temporal and spatial structure between instances can be modeled in different ways. In BoW models, this can be achieved by dividing the images \cite{Grauman2005,Lazebnik2006} or videos \cite{laptev2008} into different spatial and/or temporal zones. Each zone is characterized individually, and the final representation is the concatenation of every zone feature vectors. For audio and video, sub-sequences of instances have been analyzed using traditional sequence modeling tools such as conditional random fields (CRF) \cite{Tax2010} and hidden Markov model (HMM) \cite{guan2016}. Spatial dependency in images have also been modeled in with CRF in \cite{Warrell2011,Zha2008}.

\subsection{Data Distributions}
\label{Section:DataDistribution}
Many methods make implicit assumptions on the shape of the distributions, or on how well the negative distribution is represented by the training set. In this section, the challenges associated with the nature of the overall data distribution is studied.

\subsubsection*{Multimodal Distributions of Positive Instances}
Some MIL algorithms work under the assumption that the positive instances are located in a single cluster or region in feature space. This is the case for several early methods like APR \cite{Dietterich1997}, which searches for a hyper-rectangle that maximizes the inclusion of instances from positive bags while excluding instances from negative bags. Diverse Density (DD) \cite{Maron1998} methods follow a similar idea. These methods locate the point in feature space closest to instances in positive bags, but far from instances in negative bags. This point is considered to be the positive concept. Some more recent methods follow the single cluster assumption. CKMIL \cite{Li2014} locates the most positive instance in each bag based on its proximity to a single positive cluster center. In \cite{Xiao2016}, the classifier is a sphere encompassing at least one positive instance from each positive bag while excluding instances from negative bags. The method in \cite{Tax2010} employs a similar strategy.

\begin{figure}
\centering
\includegraphics[width=0.55\textwidth]{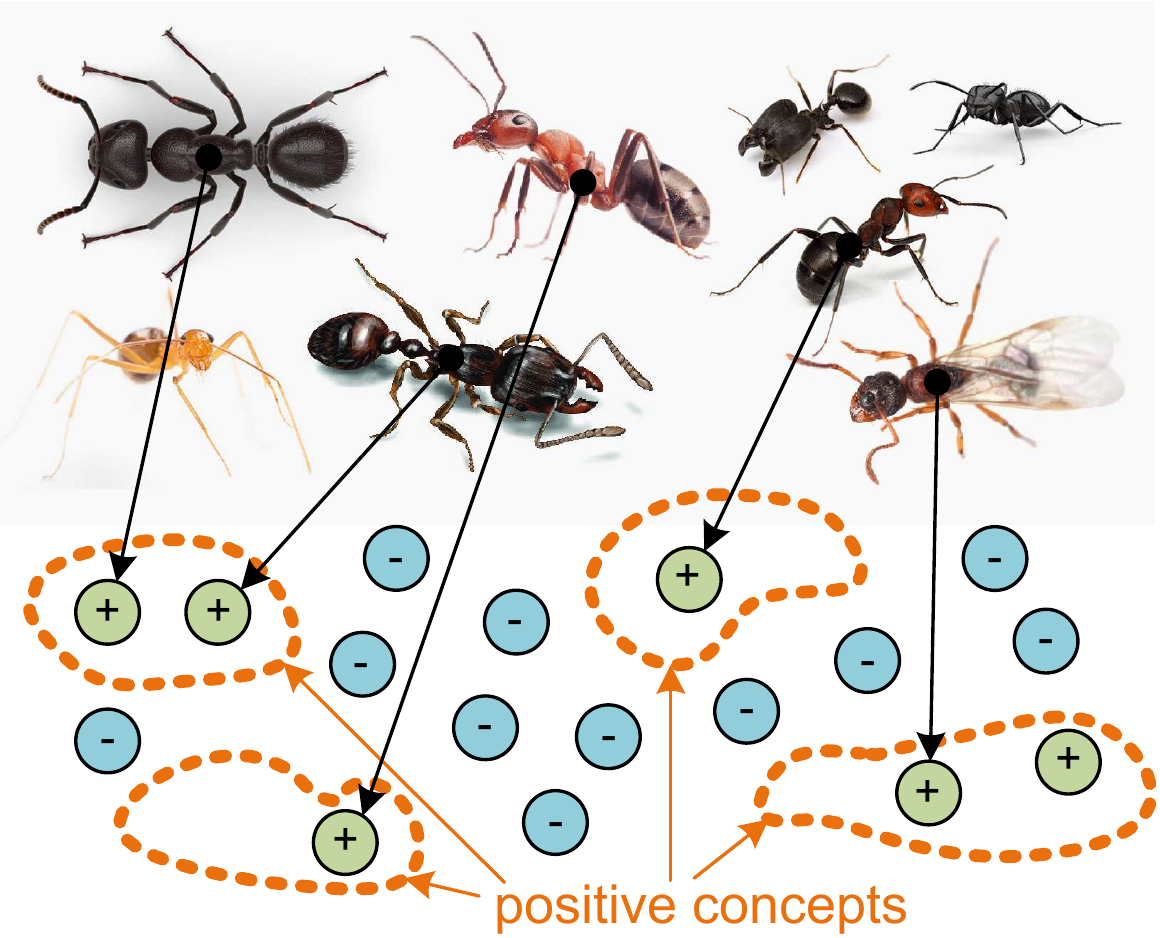}
\caption{For the same concept \textit{ants}, there can be many data clusters (modes) in feature space corresponding to different poses, colors and castes.}              
\label{Figure:multiConcept}       
\end{figure}

The single cluster assumption is reasonable in some applications such as molecule classification, but problematic in many other contexts. In image classification, the target concept may correspond to many clusters. For example, Fig. \ref{Figure:multiConcept}, shows several pictures of ants. Ants can be black, red or yellow, they can have wings and different body shapes depending on the species and castes. Their appearance also changes depending on the point-of-view. It is unlikely that a compact location in feature space encompasses all of these variations.

Many MIL methods can learn multimodal positive concepts, however, only few representative approaches will be mentioned due to space constraints. First, non-parametric methods based on distance between bags like Citation-kNN\cite{Wang2000citation} and MInD \cite{Cheplygina2015Diss} naturally deal with all shapes of distributions. Simple non-parametric methods often lead to competitive results in MIL problems \cite{Venkatesan2015}. Methods using distances to a set of prototypes as bag representation, like DD-SVM \cite{Chen2004DDSVM} and MILES \cite{Chen2006}, can model many positive clusters, because each different cluster can be represented by a different prototype. Instance-level SVM-based methods like mi-SVM \cite{Andrews02} can deal with disjoint regions of positive instances using a kernel. Also, methods modeling instance distributions in bags such as vocabulary-based \cite{Amores2010} methods naturally deal with data sets containing multiple concepts/modes. The mixture-model in \cite{Wang2012MMMIL} naturally represents different positive clusters. In \cite{Carbonneau2016} instances are grouped in clusters and the composition of the clusters are analyzed to compute the probability that instances are positive.

\subsubsection*{Non-Representative Negative Distribution}
In \cite{Doran2015These}, it is stated that learnability of instance concept requires that the distribution in test is identical to the training distribution. This is true for positive concepts, however, in some applications, the training data cannot entirely represent the negative instance distribution. For instance, provided sufficient training data, it is reasonable to expect that an algorithm learns a meaningful representation that captures the visual concept of a human person. However, since humans can be found in many different environments, ranging from jungle to spaceships, it is almost impossible to entirely model the negative class distribution. In contrast, in some applications like tumor identification in radiography, healthy tissue regions compose the negative class. These tissues possess a limited appearance range that can be modeled using a finite number of samples. 

Several methods model only the positive class, and thus are well-equipped to deal with different negative distributions in test. In most cases, these methods search for a region encompassing the positive concept. In APR \cite{Dietterich1997} the region is a hyper-rectangle, while in many others it is one, or a collection of, hyper-spheres/-ellipses \cite{Maron1998,Xiao2016,Zhang2001,Tax2008}. These methods perform classification based on the distance to a point (concept) or a region in feature space. Everything that is far enough from the point, or outside the positive region, is considered negative. Therefore, the shape of the negative distribution is unimportant. A similar argument can be made for some non-parametric methods such as Citation-kNN \cite{Wang2000citation}. These methods use the distance to positive instances, instead of positive concepts, and thus, offer the same advantage. Alternatively, the MIL problem can be seen as a one-class problem, where positive instances are the target class. Consequently, several methods using one-class SVM have been proposed \cite{Zhang2005OC,Wu2009,Wang2016onlyPos}.

Experiments in Section \ref{Section:UNI} compare reference MIL algorithms in contexts where the negative distribution is different in training and in test.

\subsection{Label Ambiguity}
\label{Section:LabelAmbiguity}
Label ambiguity is inherent to weak supervision. However, there are supplementary sources of ambiguity such as noise on labels and instance labels different from bag labels.

\subsubsection*{Label Noise}
Some MIL algorithms, especially those working under the standard MIL assumption, rely heavily on the correctness of bag labels. For instance, it was shown in \cite{Venkatesan2015} that DD is not tolerant to noise in the sense that a single negative instance in the neighborhood of the positive concept can hinder performances. A similar argument was made for APR \cite{Li2010MILD} for which a negative bag mislabeled as positive, would lead to a high FPR.

In practice, there are many situations where positive instances may be found in negative bags. There are situations where labeling errors occur, but sometimes labeling noise is inherent to the data. For example, in computer vision applications, it is difficult to guarantee that negative images contain no positive patches: An image showing a house may contain flowers, but is unlikely to be annotated as a flower image \cite{Li2015Soft}. Similar problems may arise in text classification, where a paragraph contains an analogy and thus, uses words from another subject.  

Methods working under the collective assumption can naturally deal with label noise. Positive instances found in negative bags have less impact, because these methods do not assign label solely based on the presence of a single positive instance. The methods representing bags as distributions \cite{Amores2010,Rubner2000,Doran2014dist} can naturally deal with noisy instances because a single positive instance does not significantly change the distribution of a negative bag. Methods summarizing bags by averaging the instances like NSK-kernel \cite{Gartner2002} also provide robustness to noise in a similar manner. Another strategy to deal with noise is to count the number of positive instances in bags, and establish a threshold for positive classification. This is referred as the threshold-based MI Assumption in \cite{Foulds2010}. The method proposed \cite{Li2015Soft} uses both the thresholding and the averaging strategies. The instances of a bag are ranked from most positive to less positive, and the bags are represented by the mean of the top-ranking instances and the mean of the bottom ranking instances. The averaging operation mitigates the effects of positive instance in negative bags. In \cite{Erdem2011}, robustness to label noise is obtained by using dominant sets to perform clustering and select relevant instance prototype in a bag-embedding algorithm similar to MILES \cite{Chen2006}.

\begin{figure}
\centering
\includegraphics[width=0.7\textwidth]{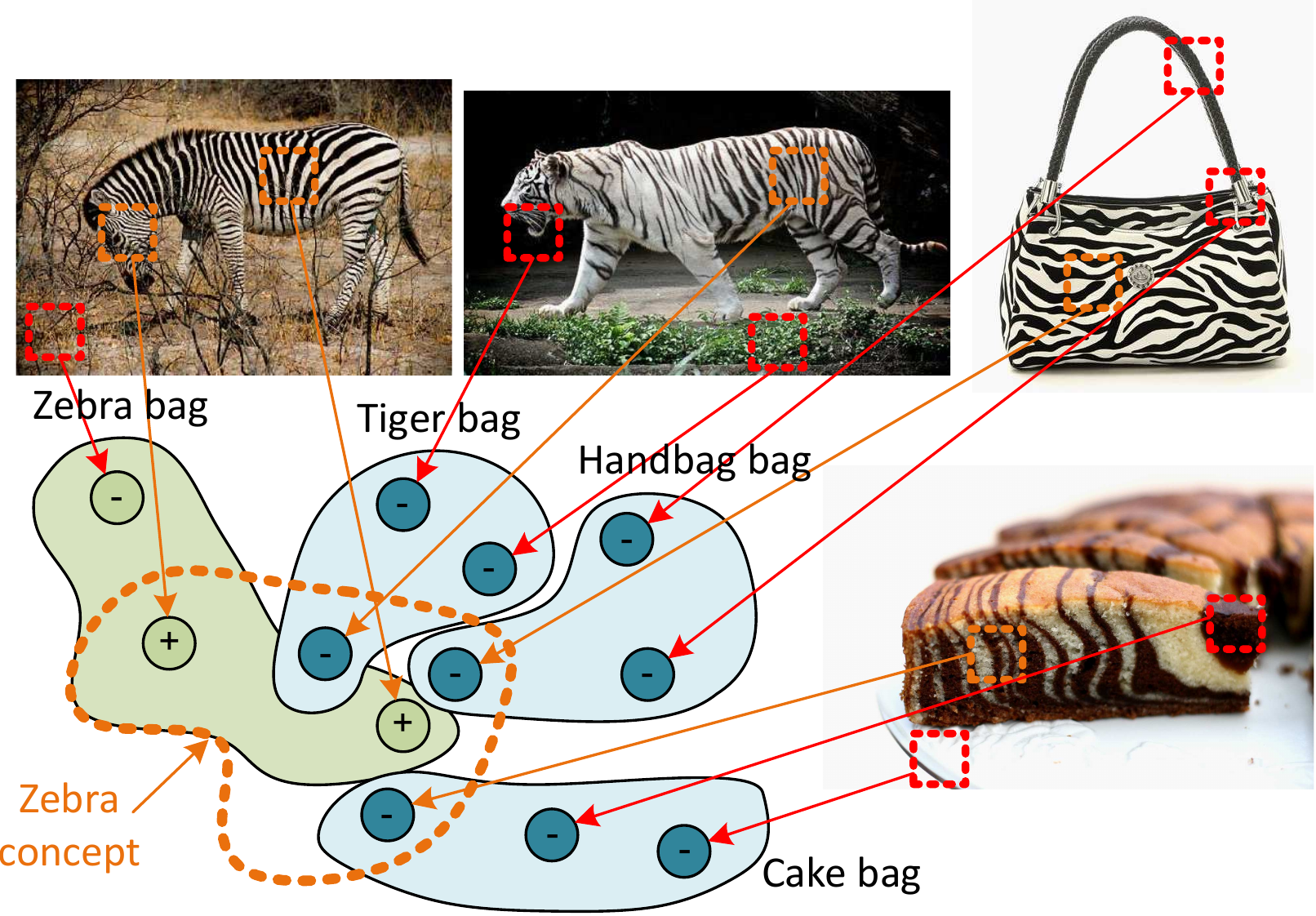}
             
\caption{This is an example of instances with ambiguous labels. \textit{Zebra} is the target concept and instances relating to this concept should fall in the region delimited by the dotted line. However, negative images can also contain instances falling inside the zebra concept region.}
\label{Figure:SoftBags}       
\end{figure}

\subsubsection*{Different Label Spaces}
There are MIL problems in which the label space for instances is different from the label space for bags. In some cases, these spaces will correspond to different granularity levels. For example, a bag labeled as a car will contain instances labeled as wheel, windshield, headlights, etc. In other cases, instances labels might not have clear semantic meanings. Fig. \ref{Figure:SoftBags} shows an example where the positive concept is zebra (represented by the region encompassed by the orange dotted line). This region contains several types of patches that can be extracted from a zebra picture. However, it is possible to extract patches from negative images that fall into this positive region. In this example, some patches extracted from the image of a white tiger, a purse and a marble cake fall into the zebra concept region. In that case the patches do not have semantic meaning easily understandable by humans. 

When instances cannot be assigned to a specific class, methods operating under the standard MIL assumption, which must identify positive instances, are inadequate. Therefore, in those cases, using the collective assumption is necessary. Vocabulary-based methods \cite{Amores2010} are particularly well adapted for this situation. They associate instances to words (e.g. prototypes or clusters) discovered from the instance distribution. Bags are represented by distributions over these words. Similarly, methods using embedding based on distance from selected prototype instance, such as MILES \cite{Chen2006} and MILIS \cite{Fu2011MILIS}, can also deal with this type of problem.

All the characteristics presented in this section define a variety of MIL problem, which each must be addressed differently. The next section relates these characteristics to the prominent application fields of MIL.

\section{Applications}
\label{Section:Applications}
MIL represents a powerful approach that is used in different application fields mostly (1) to solve problems where instances are naturally arranged in sets and (2) to leverage weakly annotated data. 



This section surveys the main application fields of MIL. Each field is examined with respect to their different problem characteristics of Section \ref{Section:keyProblems} (summarized in Table \ref{Table:matrix}). 

\begin{table} \centering

\caption{Typical problem characteristics associated with MIL in literature for different application fields (Legend: \OK ~likely to have a moderate impact, \OK\OK ~likely to have a large impact on performance)}
\scalebox{0.9}{
\begin{tabular}{l|c|c||c|c|c|c||c|c||c|c|}

& \multicolumn{10}{|c|}{\textbf{Problem Characteristics}} \\[2ex]
\textbf{Application Fields} & \rot{Instance classification} & \rot{Real-valued outputs} & \rot{Low witness rate} & \rot{Intra-bag similarities} & \rot{Instance co-occurence} 
& \rot{Structure in bags} & \rot{\shortstack[l]{Multimodal positive\\distribution}} & \rot{\shortstack[l]{Non-modelable\\ negative distribution}} 
& \rot{Label noise} & \rot{Different label spaces} \\

\cmidrule{1-11}

Drug activity prediction
& & \OK 	&	& \OK\OK &   &   & \OK & \OK &  & \\
DNA Protein identification
& \OK\OK & \OK & \OK & \OK\OK &   & \OK\OK & \OK  & \OK & &\\
Binding sites identification
& \OK\OK &\OK &  & \OK\OK &  &  & \OK & \OK &  & \\
Image Retrieval
&  &  & \OK & \OK  & \OK\OK  & \OK\OK & \OK\OK & \OK\OK & \OK & \OK\OK\\
Object localization in image
& \OK\OK &  & \OK  & \OK & \OK  & \OK & \OK\OK & \OK\OK  & \OK\OK &\OK \\
Object localization in video
& \OK\OK &  & \OK & \OK & \OK & \OK\OK & \OK\OK & \OK\OK & \OK\OK &\OK\\
Computer aided diagnosis
& \OK & \OK &\OK & \OK & \OK & & \OK & & \OK\OK &\OK \\
Text classification
& \OK &	& \OK	& 	& \OK\OK &	& \OK\OK & \OK	& \OK &\OK\\
Web mining
& \OK &	& \OK & \OK & \OK & \OK	& \OK &\OK & &\OK \\
Sound classification
& \OK &	& &\OK &\OK &\OK\OK &\OK &\OK & \OK &\\
Activity recognition
&\OK & & & &\OK &\OK\OK &\OK &\OK & \OK &\OK\\

\cmidrule[1pt]{1-11}
\end{tabular}}

\label{Table:matrix}
\end{table}

\subsection{Biology and Chemistry}

The problems in biology and chemistry can often be naturally formulated as MIL problems because of the inability to observe individual instance classes. For instance, in the molecule classification task presented in the seminal paper by Dietterich et al. \cite{Dietterich1997}, the objective is to predict if a molecule will be binding to a musk receptor. Each molecule can take many conformations, with different binding strengths. It is not possible to observe the binding strength of a single conformation, but it is possible to observe it for groups of conformations, hence the MIL problem formulation. 

Since then, MIL has found use in many drug design and biological applications. Usually, the approach is similar to Dietterich's: complex chemical or biological entities (compounds, molecules, genes, etc.) are modeled as bags. These entities are composed of parts or regions that can induce an effect of interest. The goal is to classify unknown bags and sometimes to identify witness to better understand underlying mechanisms of the biological or chemical phenomenon. MIL has been used, among others, to predict a drug's bioavailability \cite{Bergeron2012}, predict the binding affinity of peptides to major histocompatibility complex molecules \cite{Manzalawy2011}, discover binding sites governing gene expression \cite{Bandyopadhyay2015,Palachanis2014} and predict gene functions \cite{Eski2013}. 


The problems presented in this section are of various natures and it is difficult to identify key characteristics applying to all cases. However, in most cases, the bags represent many arrangements or view-points of the same entity, which translate into high intra-bag similarities. Some objects like DNA sequences produce structured bags, while the many conformations of the same molecule do not. In some problems, the objective is to identify instances responsible for an effect (e.g. drug binding). Also, many applications call for quantification, using ranking or regression, instead of classification \cite{Dooly2003} (e.g. quantifying the binding strength of a molecule), which is more difficult, or at least less documented.


\subsection{Computer Vision}
MIL is used in computer vision for two main reasons: to characterize complex visual concepts using sets of different sub-concepts, and to learn from weakly annotated data. The next subsections describe how MIL is used for content-based image retrieval (CBIR) and object localization. MIL is gaining momentum in the medical imaging community, and a subsection will also be devoted to this application field.    

\subsubsection*{Content Based Image Retrieval}
Content based image retrieval (CBIR) is probably the single most popular application of MIL. The list of publications addressing this problem is long \cite{Chen2006,Rahmani2006,Andrews02,Zhang2002CBIR,Zhang2005OC,Vijayanarasimhan2008,Maron1998Nat,Leistner2010,Song2013}. The task in CBIR is to categorize images based on the objects/concepts they contain. The exact localization of the objects is not important, which means it is primarily a bag classification problem. Typically, images are partitioned into smaller parts or segments, which are then described by feature vectors. Each segment corresponds to an instance, while the whole image corresponds to a bag. Images can be partitioned in many ways, which are compared in \cite{Wei2016}. For example, the image can be partitioned using a regular grid \cite{Maron1998Nat}, key-points \cite{Csurka2004} or semantic regions \cite{Yang2006asym,Chen2004DDSVM}. In the latter case, the images are divided using state-of-the-art segmentation algorithms. This limits instance ambiguity since segments tend to contain only one object.


This task is subject to most of the key-challenges associated with the problem characteristics in Section \ref{Section:keyProblems}. Images are a good example of non-i.i.d. data. A bag can contain many similar instances, especially if the instances are obtained using dense grid sampling. Methods using segmentation algorithms are less subject to this problem since segments tend to correspond to single objects. Some objects are more likely to co-occur in the same picture (e.g. bird and sky). Methods leveraging these co-occurrences tend to be more successful. Sometimes the subject of a picture is a composition of several concepts, which means methods working under the collective MIL assumption perform better. Working with images often means working with large intra-class variability. For instance, the same object can appear considerably different depending on the points of view. Also, many types of object can have different shapes and colors. This means it is unlikely that a unimodal distribution adequately represents the entire class. Furthermore, backgrounds can vary a lot, making it difficult to learn a negative distribution that models every possible background object. 
 

\subsubsection*{Object Localization and Segmentation}

In MIL, the localization of objects in images (or videos) means learning from bags to classify instances. Typically, MIL is used to train visual object recognition systems on weakly labeled image data sets. In other words, labels are assigned to entire images based on the objects they contain. The objects do not have to be in the foreground, and an image may contain multiple objects. In contrast, in strongly supervised applications, bounding boxes indicating the location of each object are provided along with object labels. In other cases, pixel-wise annotations are provided instead. These bounding boxes, or pixel annotations, are often manually specified, and thus, necessitate considerable human effort. The computer vision community turned to MIL to leverage the large quantity of weakly annotated images found on the Internet to build object detectors. The weak supervision can come from description sentences \cite{Xu2016Vid,Karpathy2015,Fang2015}, web search engine results \cite{Zhu2015objDisc}, tags associated with similar images and words found on web pages associated with the images \cite{Wu2015deep}. 


In several methods for object localization, bags are composed of many candidate bounding boxes corresponding to instances \cite{Hoffman2015,Song2014,Babenko2011,Babenko2008alignement,Sapienza2014}. The best bounding box to encompass the target object is assumed to be the most positive instance in the bag. Efforts were dedicated to localize objects and segment them at pixel-level using traditional segmentation algorithms such as Constraint Parametric Min-Cuts \cite{Muller2012}, JSEG \cite{Zha2008} or Multi-scale combinatorial grouping \cite{Hariharan2014}. Alternatively, segmentation can be achieved by casting each pixel of the image as an instance \cite{Vezhnevets2010}. 

Instance classification has also been applied in videos. It has been used to recognize complex events such as ``attempting a board trick'' or ``birthday party''~\cite{Phan2015,Lai2014}. Several concepts compose these complex events. Evidence of these concepts sometimes lasts only for a short time, and can be difficult to observe in the total amount of information presented in the video. To deal with this problem, video sequences are divided in shorter sequences (instances) that are later classified individually. This problem formulation is also used in \cite{Wang2011Horror} to recognize scenes that are inappropriate for children. Also in videos, MIL methods were proposed to perform object tracking \cite{Babenko2011Track,Zhang2013,Lu2011}. For example, in \cite{Babenko2011Track} a classifier is trained online to recognize and track an object of interest in a frame sequence. The tracker proposes candidate windows which compose a bag and are used to train the MIL classifier.  

It can be difficult to manually select a finite set of classes to represent every object found in a set of images. Thus, it was proposed to perform the object localization alongside class discovery \cite{Zhu2015objDisc}. The method is akin to multiple instance clustering methods \cite{Zhang2009Clust,Zhang2011M3IC}, but generates bags using a saliency detector, which remove background objects from positive bags to achieve higher cluster purity. A method based on multiple instance clustering was also proposed to discover a set of actons (sub-actions) from videos to create a mid-level representation of actions \cite{Zhu2013}.


Object localization is susceptible to the same challenges as CBIR: instances in images are correlated, exhibit high similarity and spatial (and temporal for videos) structures exist in the bags. The objects can be deformable, take various appearances and be seen from different viewpoints. This means that a single concept is often represented by a multimodal distribution, and the negative distribution cannot be entirely captured by a training set. Object localization is different from CBIR because it is an instance classification problem, which means that many bag-level algorithms are inapplicable. Also, several authors noted that in this context, MIL algorithms are sensitive to initialization \cite{Song2014,Cinbis2016}.

\subsubsection*{Computer Aided Diagnosis and Detection}

MIL is gaining popularity in medical applications. Weak labels, such as the overall diagnosis of a subject, are typically easier to obtain than strong labels, such as outlines of abnormalities in a medical scan. The MIL framework is appropriate in this situation given that patients have both abnormal and healthy regions in their medical scan, while healthy subjects have only healthy regions. The diseases and image modalities used are very diverse; applications include classification of cancer in histopathology images~\cite{xu2014weakly}, diabetes in retinal images~\cite{quellec2012multiple}, dementia in brain MR~\cite{tong2014multiple}, tuberculosis in X-ray images~\cite{melendez2014novel}, classification of a chronic lung disease in CT~\cite{cheplygina2014classification} and others.  

Like in other general computer vision tasks, there are two main goals in these applications: diagnosis (i.e. predicting labels for subjects), and detection or segmentation (i.e. predicting labels for a part of a scan). These parts can be pixels or voxels (3D pixel), an image patch or a region of interest. Different applications pursue one or both goals, and have different reasons for doing so. 

When the focus is on classifying bags, MIL classifiers benefit from using information about co-occurrence and structure of instances. For example, in~\cite{melendez2014novel}, a MIL classifier trained only with X-ray images labeled as healthy or as containing tuberculosis, outperforms its supervised version, trained on outlines of tuberculosis lesions. Similar results are observed on the task of classification of chronic obstructive pulmonary disease (COPD) from chest computed tomography images~\cite{cheplygina2014classification}. 




Literature that is focused on classifying instances is somewhat less common, which may be a consequence of the lack of instance-labeled datasets. However, the lack of instance labels is what is often the motivation for using MIL in the first place, which means instance-level evaluation is necessary if these classifiers are to be translated into clinical practice. Some papers do not perform instance-level evaluation because the classifier does not provide such output~\cite{tong2014multiple}, but state that this would be a useful extension of the method in the future. Others provide instance labels but do not have access to ground truth, thus resorting to more qualitative evaluation. For example, \cite{cheplygina2014classification} examines whether the instances classified as ``most positive'' by the classifier have similar intensity distributions to what is already known in the literature. Finally, when instance-level labels are available, the classifier can be evaluated quantitatively and/or qualitatively. Quantitative evaluation is performed in~\cite{Kandemir2015,melendez2014novel,quellec2012multiple}. In addition, the output of the classifier can be displayed in the image, which is an interpretable way of visualizing the results. In~\cite{melendez2014novel}, the mi-SVM classifier provides local real-valued tuberculosis abnormality scores for each pixel in the image, which are then visualized as a heatmap on top of the X-ray image.

Like other computer vision tasks, CAD is subject to most of the key-challenges discussed in Section \ref{Section:keyProblems}. Depending on the sampling - which can be done on a densely-sampled grid~\cite{Kandemir2015,melendez2014novel}, randomly~\cite{cheplygina2014classification}, or according to constraints \cite{tong2014multiple} -- the instances can display varying degrees of similarity. In many pathologies, abnormalities are likely to include different subtypes, which have different appearance resulting in multimodal concept distributions. Moreover, differences between patients, such as age, sex and weight, as well as differences in acquisition of the images also can lead to large intra-class variability. On the other hand, the negative distribution (healthy tissue) is more constrained than in computer vision applications. CAD problems are naturally suitable to have real-valued outputs, because diseases can have different stages, although this is often not considered when off-the-shelf algorithms are applied. For example, the chronic lung disease COPD has 4 different stages, but~\cite{cheplygina2014classification} treats them all as the positive class. During evaluation, the mild stage is most often misclassified as healthy. \cite{tong2014multiple} considers binary classification tasks out of four possible classes (healthy, two types of mild cognitive impairment, and Alzheimer's), while these could be considered as a continuous scale. Lastly, as for other applications, the difference between bag-level and instance-level classification presents an important challenge. 





\subsection{Document Classification and Web Mining}
Considering the Bag-of-Words (BoW) model is a MIL model working under the collective assumption, document classification is one of the earliest (1954) applications of MIL \cite{harris1954distributional}. BoW represents texts as frequency histograms quantifying the occurrence of each word in the text. In this context, texts and web pages are multi-part entities that require MIL classification framework. 

Texts often contain several topics and are easily modeled as bags. Text classification problems can be formulated as MIL at different levels. At the lowest level, instances are words like in the BoW model. Alternatively, instances can be sentences \cite{Zhang2008advertisement,Pappas2014}, passages \cite{Andrews02,Zhang2013MI2LS} or paragraphs \cite{Ray2005}. In \cite{Andrews02}, bags are text documents, which are divided in overlapping passages corresponding to instances. The passages are represented by a binary vector in which each element is a medical term. The task is to categorize the texts. In \cite{Settles2008}, instances are short posts from different newsgroups. A bag is a collection of posts and the task is to determine if a group of posts contains a reference to a subject of interest. In \cite{Ray2005}, the task consists of identifying texts that contain a passage which links a protein to a particular component, process or function. In this case, paragraphs are instances while entire texts are bags. The paragraphs are represented by a BoW alongside distances from the protein names and key terms. In \cite{Jorgensen2008}, the content of emails is analyzed to detect spam. A common approach to elude spam filters is to include words that are not associated with spam in the message. Representing emails as bags of passages proved to be an efficient way to deal with these attacks. In \cite{Kotzias2014,Kotzias2015kdd,Zhang2008advertisement,Pappas2014}, MIL was used to infer the sentiment expressed in individual sentences based on the labels provided for entire user reviews. MIL has also been used to discover relations between named entities \cite{Bunescu2007relation}. In this case, bags are collections of sentences containing two words that may or may not express a target relation (e.g. "Rick Astley" lives in "Montreal"). If the two words are related in the specified way, some of the sentences in the bag will express this relation. If that is not the case, none of the sentences will indicate the relation, hence the MIL formulation. 

Web pages can also be naturally modeled using the MIL framework. Just like texts, web pages often contain many topics. For instance, a news channel website contains several articles on a diversity of subjects. MIL has been used for web index-page recommendations based on a user browsing history \cite{Zhou2005web,Zafra2007}. A web index page contains links, titles and sometimes short description of web pages. In this context, a web index page is a bag, and the linked web pages are the instances. Following the standard MIL assumption, it is hypothesized that if a web index page is marked as favorite, the user is interested in a least one of the pages linked to it. Web pages are represented by the set of the most frequent terms they contain. In contextual web advertisement, advertisers prefer to avoid certain pages containing sensitive content like war or pornography. In \cite{Zhang2008advertisement}, a MIL classifier assesses sections of web pages to identify suitable web pages for advertisement.  

The classification of web and text documents is subject to most of the difficulties associated with MIL problem characteristics. Depending on the task and the formulation of the problem, bag and instance classification can be performed. Often only small passages or specific words indicate the class of the document, which means WR can be quite low. Words may have different meanings depending on the context and thus, co-occurrence is important in this type of application. While structure is an important component of sentences, most of the existing MIL methods discard it. In addition, text classification can present an additional difficulty compared to other applications. When texts are represented by a BoW the data is very sparse and high-dimensional \cite{Andrews02}. This type of data is often difficult to handle by classifiers using Euclidean-like distance measures. These distributions are highly multimodal and it is difficult to adequately represent the negative distribution.

\subsection{Other Applications}
The MIL formulation has found its way to various other application fields. In this section, we present some less common applications for MIL along with their respective formulation.

Reinforcement learning (RL) shares some similarities with MIL. In both cases, only a weak supervision is provided for the instances. In RL, a reward, the weak supervision, is assigned to a state/action pair. The reward obtained for the state/action pair is not necessarily directly related to it, but might be related to preceding actions and states. Consider a RL agent learning how to play chess. The agent obtains a reward (or punishment) only at the end of the game. In other words, a label is given for a collection (bag) of action/state pairs (instances). This correspondence has motivated the use of MIL to accelerate RL by the discovery of sub-goals in a task \cite{Mcgovern2003}. These sub-goals are, in fact, the positive instances in the successful episodes. The main challenge for RL task is to consider the structure in bags and the label noise since good actions can be found in bad sequences.

Just like for images, some sound classification tasks can be cast as MIL. In \cite{Mandel2008}, the objective is to automatically determine the genre of musical excerpts. In training, labels are provided for entire albums or artists, but not for each excerpt. The bags are collection of excerpts from the same artist or album. It is possible to find different genres of music on the same album or from the same artist, therefore the bags may contain positive and negative instances. In \cite{Briggs2012}, MIL is used to identify bird songs in recordings made by an unattended microphone in the wild. Sound sequences contain several types of birds and other noises. The objective is to identify each birdsong individually while training only on weakly labeled sound files.

Some methods represent audio signals as spectrograms and use image recognition techniques to perform recognition \cite{Lyon2010}. This idea has been used for bird song recognition \cite{Ruiz2015} with histograms of gradients. In \cite{Carbonneau2016Personality}, personality traits are inferred from speech signals represented as spectrograms in a BoW framework. In that case, entire speech signals are bags and small parts of the spectrogram are instances. The BoW framework has been used in a similar fashion in \cite{Kumar2016}, however, in that case instances are cepstrum feature vectors representing 1 second-long audio segments. In general, audio classification is subject to the same challenges as image classification applications.

Time series are found in several applications other than audio classification. For instance, in \cite{Stikic2011,guan2016} MIL is used to recognize human activities from wearable body sensors. The weak supervision comes from the users stating which activities were performed in a given time period. Typically, activities do not span across entire periods and each period may contain different activities. In this setup, instances are sub-periods, while the entire periods are bags. A similar model is used for the prediction of hard drive failure \cite{Murray2005}. In this case, time series are a set of measurements on hard drives taken at regular intervals. The goal is to predict when a product is about to fail. Time series imply structure in bags that should not be ignored. 

In \cite{Manandhar2012,Karem2011}, MIL classifiers detect buried landmines from ground-penetrating radar signals. When a detection occurs at a given GPS coordinate, measures are taken at various depths in the soil. Each detection location is a bag containing feature vectors for different depths. 

In \cite{Maron1998}, MIL is used to select stocks. Positive bags are created by pooling the 100 best-performing stocks each month, while negative bags contain the 5 worst performing stocks. An instance classifier selects the best stocks based on these bags.  

In \cite{Mcgovern2003}, a method learning relational structure in data predicts which movies will be nominated for an award. A movie is represented by a graph that models its relations to actors, studios, genre, release date, etc. The MIL algorithm identifies which sub-graph explains the nomination to infer the success of test cases. This type of structural relation between bags and instance is akin to web page classification problems.


\section{Experiments}
\label{Section:Experiments}

In this section, 16 reference methods are compared using data sets that allows to shed in light on some of the problem characteristics discussed in Section \ref{Section:keyProblems}. These experiments are conducted to show how problem characteristics influence the behavior of MIL algorithms, and demonstrate that these characteristics cannot be neglected when designing or comparing MIL algorithms. Three characteristics were selected, each from a different category, to represent the spectrum of characteristics. Algorithms are compared on the instance classification task, under different WR and with an unobservable negative distribution. These characteristics were chosen because their effect can be isolated and easily parametrized. The reference methods used in the experiments were chosen because they represent a most families of approaches and include most of the most widely used reference methods. 

\subsection{Reference Methods}
\label{Section:ReferenceMethods}

\subsubsection*{Instance Space Methods}

\textbf{SI-SVM, SI-SVM-TH and SI-\textit{k}NN: }
These are not a MIL method \textit{per se}, but give an indication on the pertinence of using MIL methods instead of regular supervised algorithms. In these algorithms, each instance is assigned the label of its bag, and bag information is discarded. The classifier assign a label to each instance, and a bag is positive if it contains at least one positive instance. For SI-SVM-TH the number of positive instances detected is compared to a threshold that is optimized on the training data. 

\textbf{MI-SVM and mi-SVM \cite{Andrews02}:}
These algorithms are transductive SVMs. Instances inherit their bag label. The SVM is trained and classify each instance in the data set. It is then retrained using the new label assignments. This procedure is repeated until the labels remain stable. The resulting classifier is used to classify test instances. MI-SVM uses only the most positive instance of each bag for training, while mi-SVM uses all instances.

\textbf{EM-DD \cite{Zhang2001}:}
DD \cite{Maron1998} measure the probability that a point in feature space belongs to the positive class given the class proportion of instances in the neighborhood. EM-DD uses the Expectation-Maximization algorithm locate the maximum of the DD function. Classification is based on the distance from this maximum point.

\textbf{RSIS \cite{Carbonneau2016}:}
This method probabilistically identifies the witnesses in positive bags using a procedure based on random subspacing and clustering introduced in \cite{Carbonneau2016ICPR}.  Training subsets are sampled using the probabilistic labels of the instance to train an ensemble of SVM.

\textbf{MIL-Boost \cite{Babenko2008alignement}:}
The MIL-Boost algorithm used in this paper is a generalization of the algorithm presented in \cite{Viola2006MIL}. The method is essentially the same as gradient boosting \cite{Friedman2001} except that the loss function is based on bag classification error. The instances are classified individually, and their labels are combined to obtain bag labels.

\subsubsection*{Bag Space Methods}

\textbf{C-kNN \cite{Wang2000citation}:} This is an adaptation of kNN to MIL problems. The distance between two bags is measured using the minimal Hausdorff distance. C-kNN relies on a two-level voting scheme inspired from the notion of citations and references in research papers. The algorithm was adapted in \cite{Zhou2005cknnroi} to perform instance classification. 

\textbf{MInD \cite{Cheplygina2015Diss}:} With this method, each bag is encoded by a vector whose fields are dissimilarities to the other bags in the training data set. A regular supervised classifier, an SVM in this case, classifies these feature vectors. Many dissimilarity measures are proposed in the paper, but the \textit{meanmin} offered the best overall performance and will be used in this paper.
 
\textbf{CCE \cite{Zhou2007CCE}:} This algorithm is based on clustering and classifier ensembles. At first, the feature space is clustered using a fixed number of clusters. The bags are represented as binary vectors in which each bit corresponds to a cluster. A bit is set to 1 when at least one instance in a bag is assigned to its cluster. The binary codes are used to train one of the classifiers in the ensemble. Diversity is created in the ensemble by using a different number of clusters each time. 

\textbf{MILES \cite{Chen2006}:} In Multiple-Instance Learning via Embedded instance Selection (MILES) an SVM classifies bags represented by a feature vectors containing maximal similarities to selected prototypes. The prototypes are instances from the training data selected by a 1-norm SVM. Instance classification relies on a score representing the instance contribution to the bag label.

\textbf{NSK-SVM \cite{Gartner2002}:} The normalized set kernel (NSK) basically averages the distances between all instances contained in two bags. The kernel is used in an SVM framework to perform bag classification.

\textbf{mi-Graph \cite{Zhou2009migraph}:} This method represents each bag by a graph in which instances correspond to nodes. Cliques are identified in the graph to adjust the instances weights. Instances belonging to large cliques have lower weight so that every concept present in the bag is equally represented when instances are averaged. A graph kernel captures similarity between bags and is used in an SVM.

\textbf{BoW-SVM:} Creating a dictionary of representative words is the first step when using a BoW method. This is achieved with BoW-SVM by performing k-means clustering on all the training instances. Next, instances are represented by the most similar word contained in the dictionary. Bags are represented by frequency histograms of the words. Histograms are classified by an SVM using a kernel suitable for histogram comparison (exponential $\chi ^2$ in this case).

\textbf{EMD-SVM}:
The Earth Mover distance (EMD) \cite{Rubner2000} is a measure of the dissimilarity between two distributions. Each bag is a distribution of instances and the EMD is used to create a kernel used in an SVM.

\subsection{Data Sets}
\label{Section:DataSets}

\textbf{Spatially Independent, Variable Area, and Lighting (SIVAL) \cite{Rahmani2005}:} This data set contains 500 images each segmented and manually labeled by \cite{Settles2008}. It contains 25 classes of complex objects photographed from different viewpoints in various environments. Each bag is an image partitioned in approximately 30 segments. A 30-dimensional feature vector encodes the color, texture and neighbor information of each segment. There are 60 images in each class, which are in turn considered as the positive class. 5 randomly selected images from each of the 24 other classes yield 120 negative bags. The data sets are generated 5 times. The WR is 25.5\% in average but ranges from 3.1 to 90.6\%. In this data set, unlike in other image data sets, co-occurrence information between the objects of interest and the background is nonexistent because all 25 objects are photographed in the same environment. 

\textbf{Birds \cite{Briggs2012}:} The bags of this data set correspond to 10 seconds recordings of bird songs from one or more species. The recording is segmented temporally to create instances, which belong to a particular bird or to background noises. These 10232 instances are represented by 38-dimensional feature vectors. Readers should refer to the original paper for details on the features. There are 13 types of bird in the data set, each in turn considered as the positive class. Therefore 13 problems are generated from this data set. In this data set, low WR poses a challenge, especially since it is not constant across bags. Moreover, bag classes are sometimes severely imbalanced.

\textbf{Newsgroups \cite{Settles2008}:} The newsgroups data set was derived from the \textit{20 Newsgroups} \cite{Lang1995} data set corpus. It contains posts from newsgroups on 20 subjects. Each post is represented by 200-term frequency-inverse document frequency (TFIDF) features. This representation generally yields sparse vectors, in which each element is representative of a word frequency in the text scaled by its frequency in the entire corpus. When one of the subjects is selected as the positive class, all 19 other subjects are used as the negative class. The bags are collections of posts from different subjects. The positive bags contain an average of 3.7\% of positive instances. This problem is semi-synthetic and does not correspond to a real-world application. There is thus no exploitable co-occurrence information, intra-bag similarities or bag structure. However, the representation yields sparse data, which is different from the two previous data sets, and is representative of text applications.

\textbf{HEPMASS \cite{Baldi2016}:}
The instances of this data set come from the HEPMASS Data Set\footnote{\url{http://archive.ics.uci.edu/ml/datasets/HEPMASS}}. It contains more than 10M instances which are simulation of particle collisions. The positive class correspond to collisions that produce exotic particles, while the negative class is background noise. Each instance is represented by a 27-dimensional feature vector containing low-level kinematic measurements and their combination to create higher level mass features (see original paper for more details). For each WR value, 10 versions of the MIL data are randomly generated. For each version, the training and a test sets contain 50 positive bags and 50 negative bags composed of 100 instances. 

\textbf{Letters \cite{Frey1991}:}
This semi-synthetic MIL data set uses instances from the Letter Recognition data set\footnote{\url{https://archive.ics.uci.edu/ml/datasets/Letter+Recognition}}. It contains a total of 20k instances representing each of the 26 letters in the English alphabet. Each of these letters can be seen as a concept and used to create different positive and negative distributions. Each letter is encoded by a 16-dimensional feature vector that has been standardized. The reader is referred to the original paper for more details. In WR experiments, for each WR value, 10 versions of the MIL data sets are randomly generated. Each version has a training and a test set. Both sets contain 50 positive bags and 50 negative bags each containing 20 instances. In the positive bags, witness are sampled from 3 letters randomly selected to represent positive concepts. All other letters are considered are negative concepts. For the experiments on negative class modeling, the data set is divided in train and test partitions each containing 200 bags. Each bag contains 20 instances. The bag classes are equally proportioned and the WR is 20\%. Like before, the positive instances are samples from 3 randomly selected letters. Half of the remaining letters constitute the initial negative distribution and the other half constitutes the unknown negative distribution. 

\textbf{Gaussian Toy Data:} In this synthetic data set, the positive instances are drawn from a 20-dimensional multivariate Gaussian distribution ($\mathcal{G}(\mathbf{\mu},\mathbf{\Sigma})$) that represents the positive concept. The values of $\mathbf{\mu}$ are drawn from $\mathcal{U}(-3,3)$. The covariance matrix ($\mathbf{\Sigma}$) is a randomly generated semi-definite positive matrix in which the diagonal values are scaled to $\left ] 0,0.1 \right ]$. The negative instances are sampled from a randomly generated mixture of 10 similar Gaussian distributions. This distribution is gradually replaced by another randomly generated mixture. The data set is standardized after generation. The test and training partitions both contain 100 bags. There are 20 instances in each bag and the WR is 20\%.

\subsection{Instance-Level Classification}
\label{Section:IC}
\label{Section:ExpIC}

In this section, the reference methods with instance classification capabilities will be compared on three benchmark data sets: SIVAL, Birds and Newsgroups. These data sets are selected because they represent three different application fields and because instance labels are provided, which is somewhat uncommon with MIL benchmark data sets. There already exist several comparative studies for bag-level classification, we refer interested reader to \cite{Amores2013,Kandemir2015}.

The experiments were conducted using a nested cross-fold validation protocol \cite{stone1974cross}. It consists of two cross-validation loops. An outer loop assesses the performance of the algorithm in test, and an inner loop is used to optimize the algorithm hyper-parameters. This means that for each test fold of the outer loop, hyper-parameters optimization is performed via grid-search. Average performance is reported on results for the outer loop test folds.

\begin{figure}[!ht]
\centering
\includegraphics[width=8.2cm]{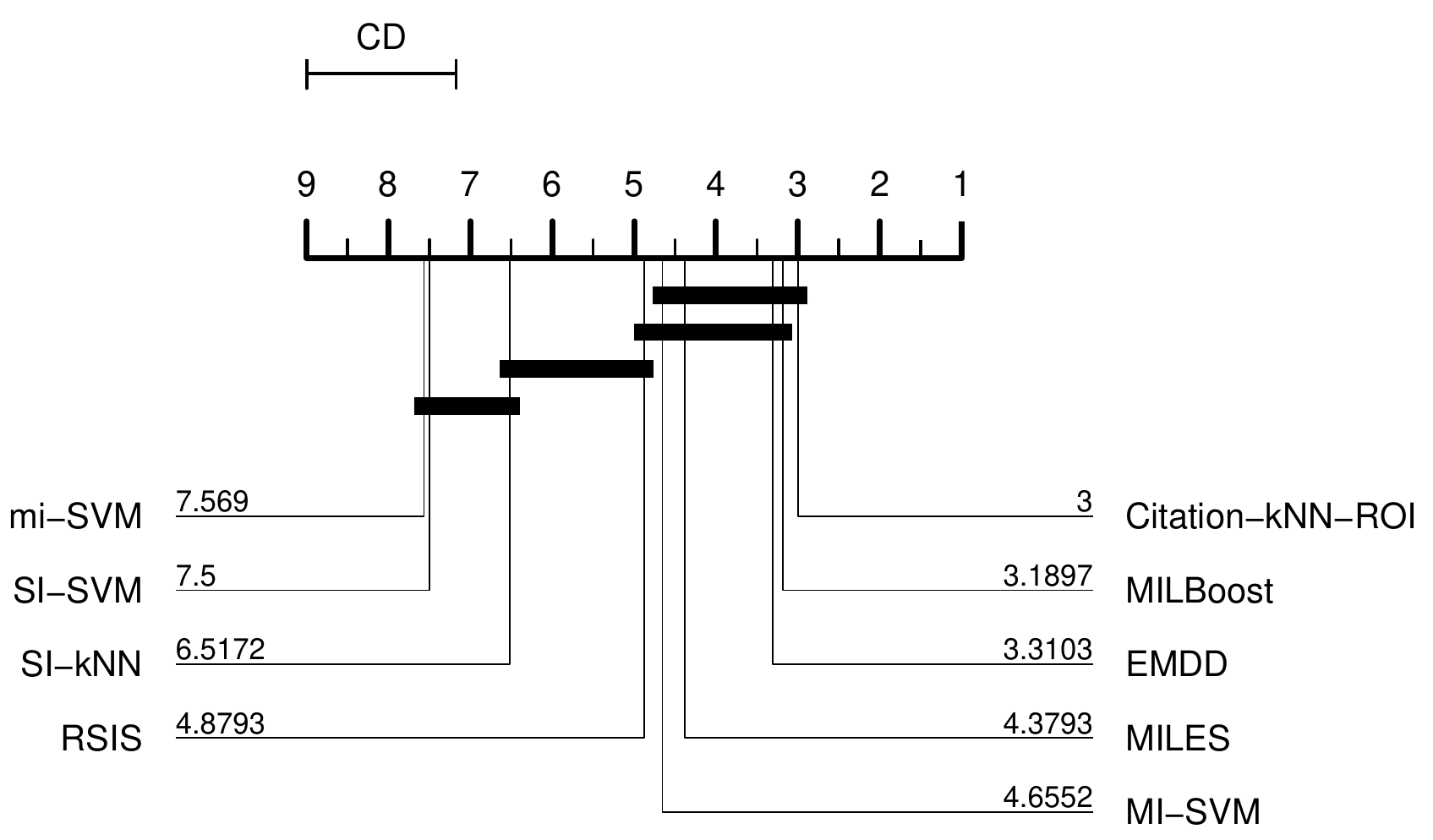}
\caption{Critical difference diagram for UAR on instance classification ($\alpha = 0.01$). Higher numbers are better.}
\label{Figure:CDUAR}
\end{figure}

\begin{figure}[!ht]
\centering
\includegraphics[width=8.2cm]{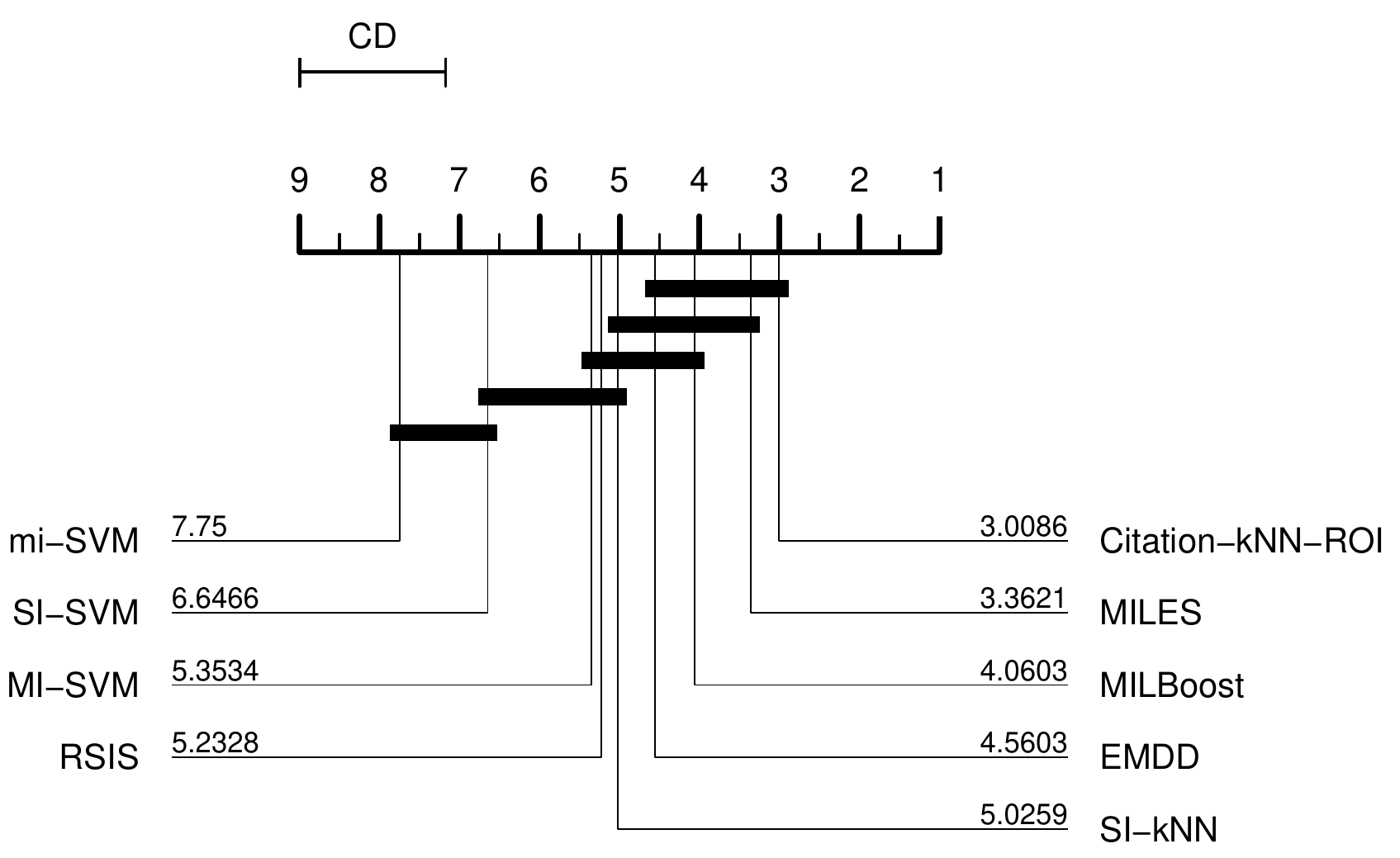}
\caption{Critical difference diagram for the $F_1$-score on instance classification ($\alpha = 0.01$). Higher numbers are better.}
\label{Figure:CDF1}
\end{figure}

Instance classification problems often exhibit class imbalance, especially when the WR is small. In these cases, comparing algorithm is terms of accuracy can be misleading. In this section, algorithms are compared in terms of unweighted average recall (UAR) and $F_1$-score. The UAR is the average of the accuracy for each class. The $F_1$-score is the harmonic mean between precision and recall. The 3 data sets translate into 58 different problems. For easy comparison, Fig. \ref{Figure:CDUAR} and \ref{Figure:CDF1} present the results in the form of critical difference diagrams \cite{Demsar2006} with a significance level of 1\%. 

\textbf{Results indicate that a successful strategy for instance classification is to discard bag information}. With both metrics, the best algorithms are mi-SVM and SI-SVM, which assign the bag label to each instance and then treat them as atomic elements. This is consistent to the results obtained in \cite{Kandemir2015}. These two methods are closely related because SI-SVM corresponds to the first iteration of mi-SVM. SI-kNN also yield competitive results and uses the same strategy. Even if the Birds and the Newsgroups data sets both possess low WR, it would seem that supervised methods are better suited for this task than MIL methods which use bag accuracy as an optimization objective (MILES, EMDD and MIL Boost). MI-SVM and RSIS rely on the identification of the most positive instances in each bag. This strategy seems successful to some degree, but is prone to ignore more ambiguous positive instances that are dominated by the others in the same bag. These conclusions have also been observed in the results obtained on the individual data sets.

\subsection{Bag Composition: Witness Rate}
\label{Section:WRexp}

These experiments study the effects of the WR on MIL algorithm performances. Two semi-synthetic data sets were created to allow control over the WR, and observe the behavior of the reference methods in greater detail: Letters and HEPMASS. These data sets are created from supervised problems that were artificially arranged in bags. This has the advantage of eliminating any structure and co-occurrence in the data, and thus better isolate the effect of WR. The original data sets must possess a high number of instances to emulate low WR. In the Letters data set, the positive class contains three concepts while in HEPMASS there is only one concept, which has an impact for some algorithms.

\begin{figure}
\centering
\includegraphics[width=0.75\textwidth]{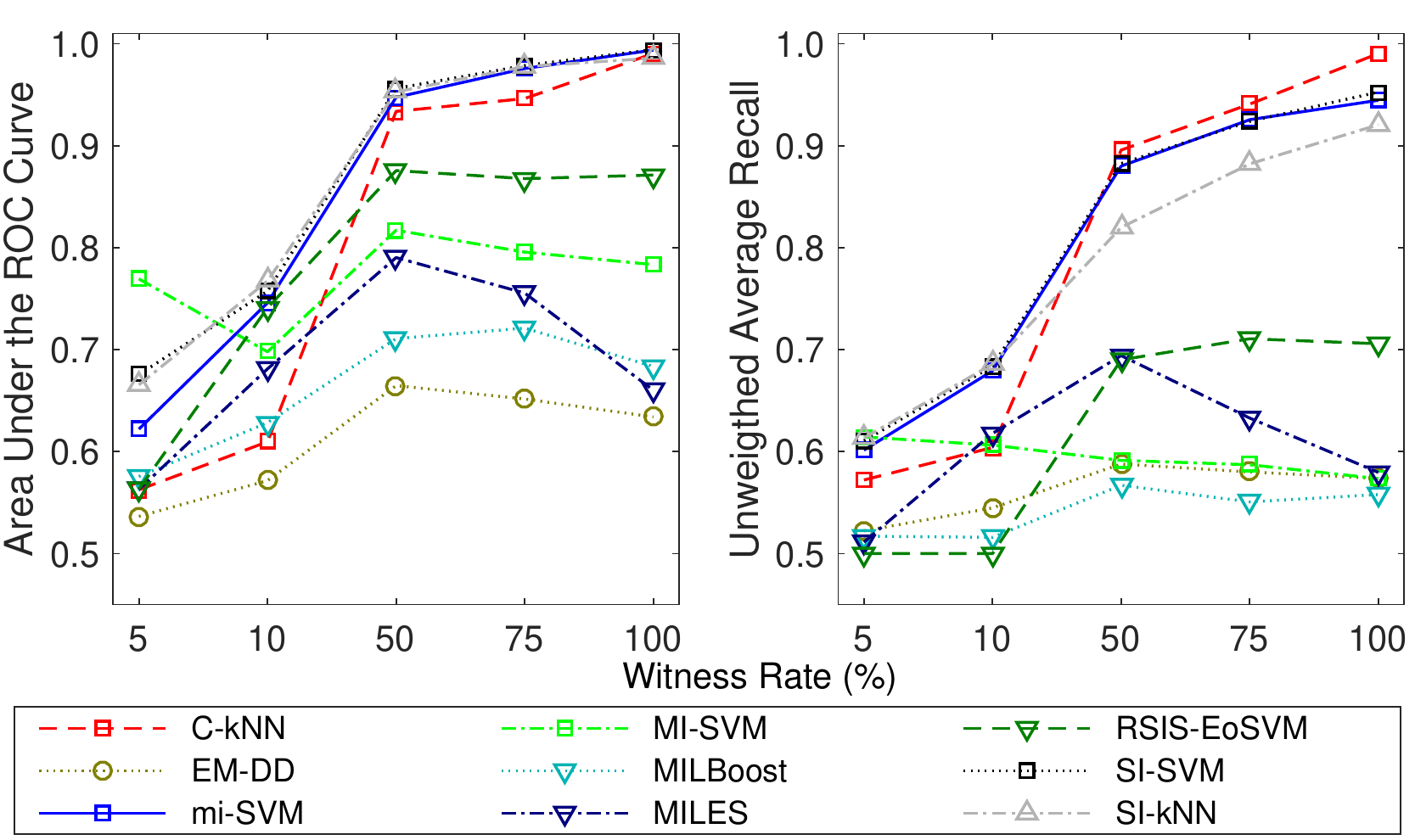}
\caption{Average performance of the MIL algorithms for instance classification on the Letters data set as the witness rate increases.}
\label{fig:WRLI}       

\centering
\includegraphics[width=0.75\textwidth]{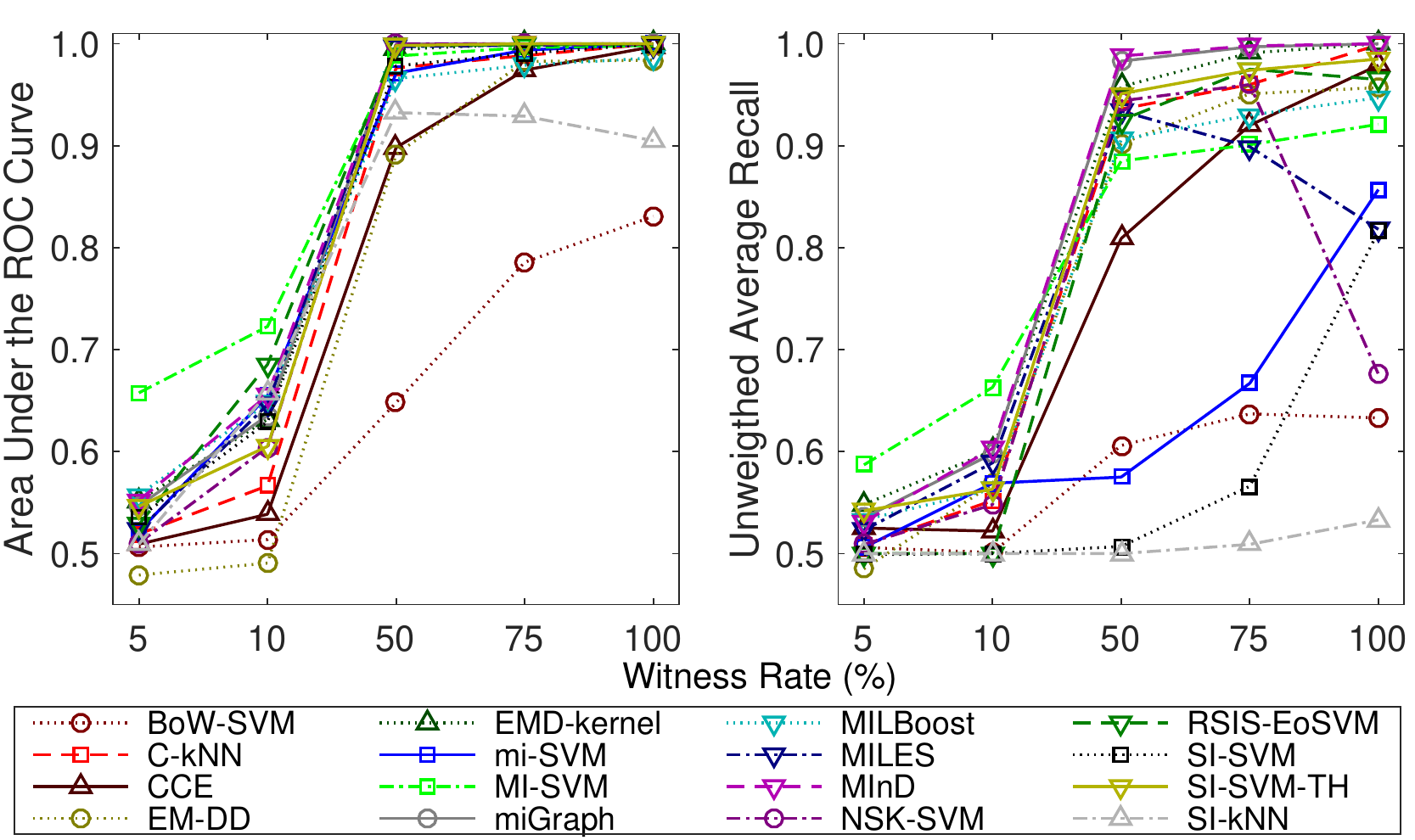}
\caption{Average performance of the MIL algorithms for bag classification on the Letters data set as the witness rate increases.}
\label{fig:WRLB}       
\end{figure}

\begin{figure}
\centering
\includegraphics[width=0.75\textwidth]{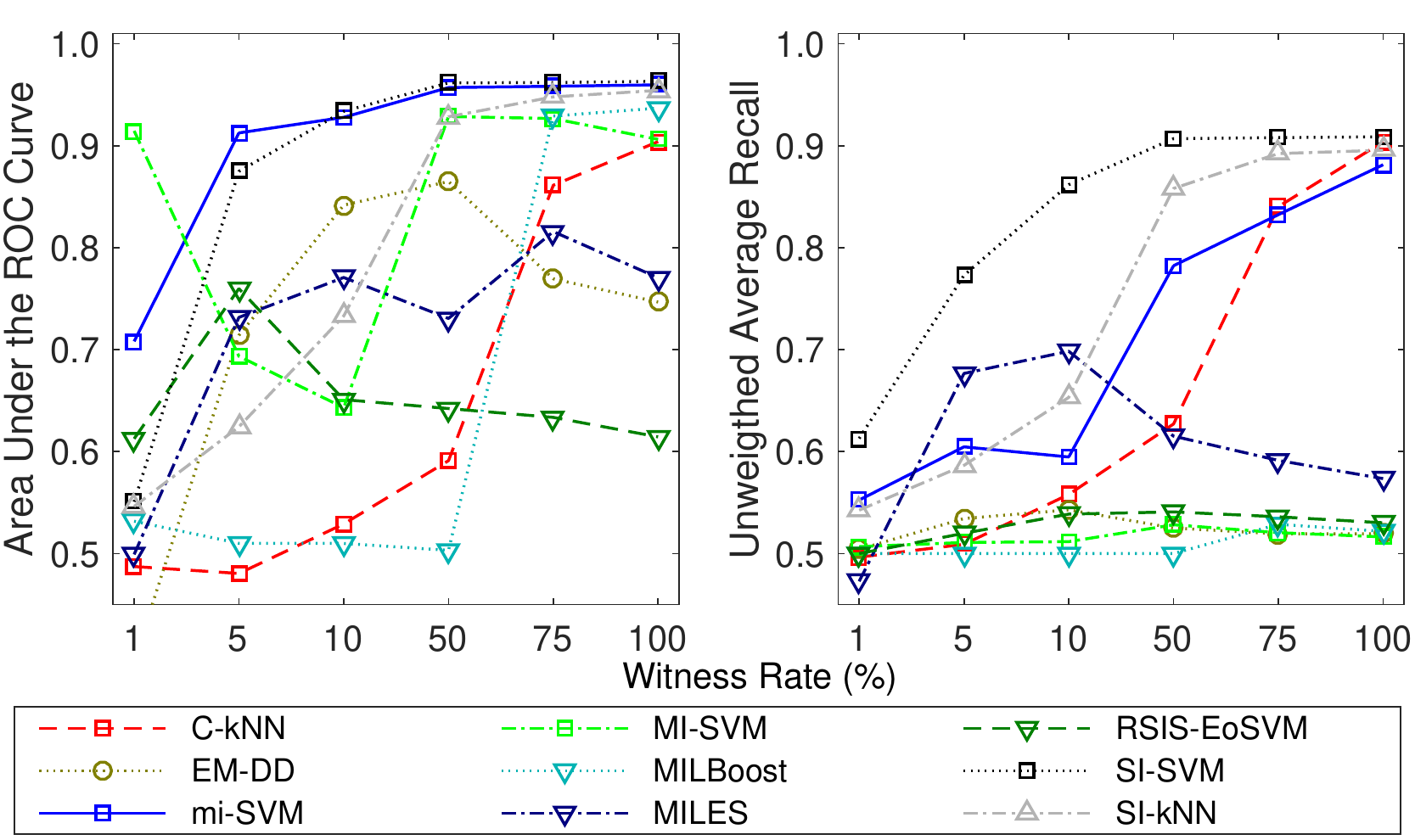}
\caption{Average performance of the MIL algorithms for instance classification on the HEPMASS data set as the witness rate increases.}
\label{fig:WRPI}       

\centering
\includegraphics[width=0.75\textwidth]{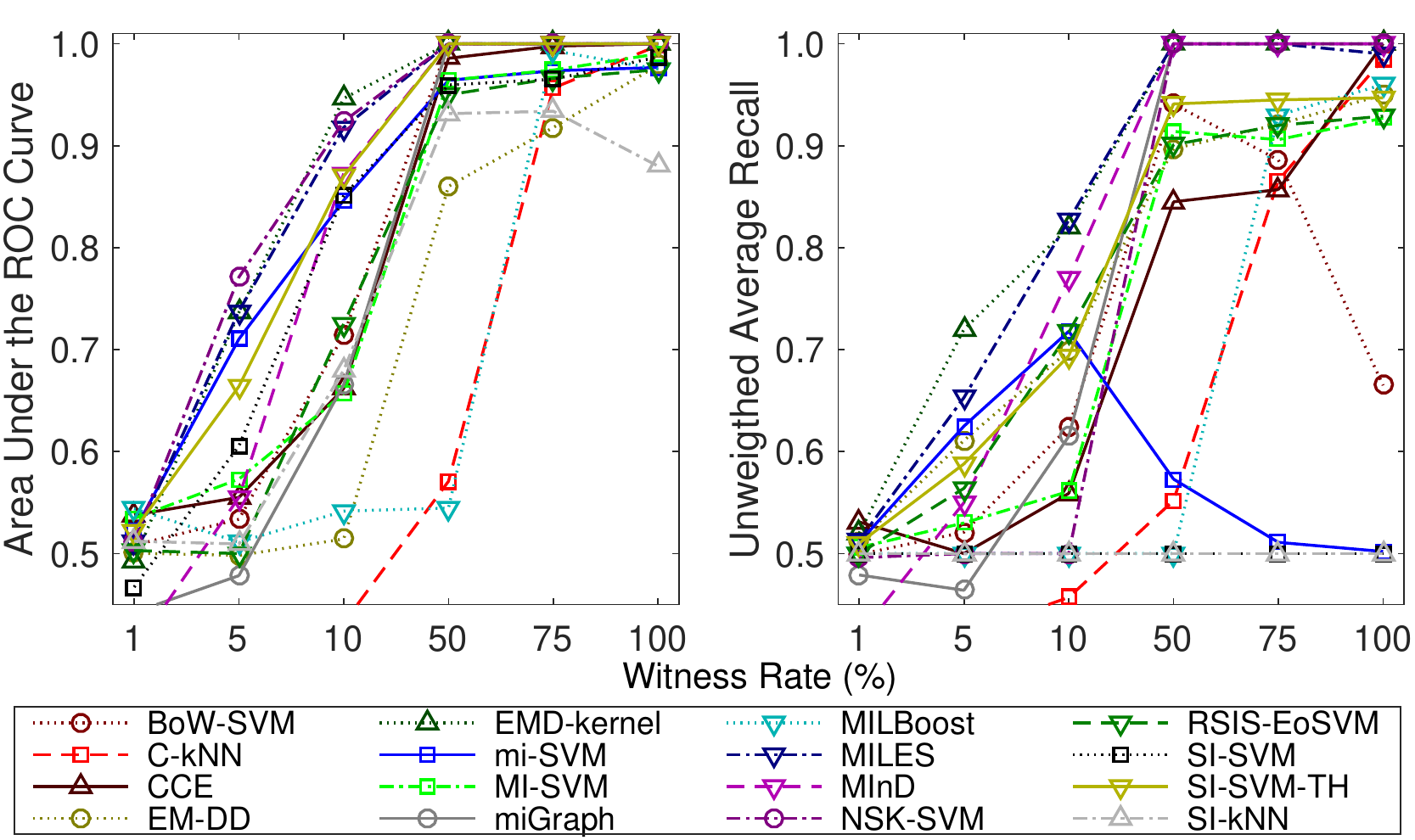}
\caption{Average performance of the MIL algorithms for bag classification on the HEPMASS data set as the witness rate increases.}
\label{fig:WRPB}       
\end{figure}

All hyper-parameters were optimized for each version of the data sets, and for each WR value using grid search and cross-validation. The results reported in Fig. \ref{fig:WRLI}, \ref{fig:WRLB}, \ref{fig:WRPI} and \ref{fig:WRPB} are the average results obtained on the test data for each of the 10 generated versions. Performance are compared using AUC and the UAR.

There are several things that can be concluded by examining the experiment results. Firstly, \textbf{for all methods, lower WR translates into lower accuracy}. However, Fig. \ref{fig:WRLI} shows that \textbf{for the instance classification task, higher WR does not necessarily means higher accuracy} for all methods. In fact, for the Letters data set, three different letters are used to create positive instances which makes the positive distribution multimodal. As discussed in Section \ref{Section:IC}, some methods are optimized for bag classification (EM-DD, MI-SVM, MILES, MILBoost, RSIS-EoSVM). In those cases, once a letter is assigned to the positive class in a positive bag, the bag is correctly classified. The remaining positive letters can be ignored and the algorithm still achieves perfect bag classification. This can be observed by comparing Fig. \ref{fig:WRLI} and \ref{fig:WRPI} with Fig. \ref{fig:WRLB} and \ref{fig:WRPB}, where the methods optimized for bag classification deliver lower accuracy for instance classification, but their accuracy is comparable to other instance-based methods when classifying bags. This explains in part the observation \cite{Vanwinckelen2015,Doran2014} that an algorithm performance for one task is not always representative of the performance in the other.

The results in Fig. \ref{fig:WRLI} and \ref{fig:WRPI} suggest that \textbf{supervised classifiers are as effective for instance classification as the best MIL classifiers when the WR is over 50\%}. In this case, the mislabeled negative instance are just noise in the training set, which easily is dealt with by the SVM or the voting scheme of the SI-kNN. Even when WR is lower than 50\% supervised methods perform better than some of their MIL counterparts. MI-SVM has higher AUC performance when the WR is at its lowest compared to the other method. This is explained by the fact that positive bags are represented by their single most positive instance. When the WR is at its minimum, there is only one witness per bag which coincides with this representation.  


\begin{table}[h]
\centering
\caption{Ranking of instance-based methods \textit{vs}. bag-based methods for the bag classification task.}
\label{my-label}
\scalebox{0.9}{
\begin{tabular}{@{}llcc@{}}
& & \multicolumn{2}{c}{\textbf{WR}} \\
\cmidrule{3-4} 
\textbf{Metric} & \textbf{Method type} 
&$<50\%$&$\geq50\%$\\
\midrule
\multirow{2}{*}{\begin{tabular}[c]{@{}l@{}}Mean rank \\ (AUC)\end{tabular}} & Instance-based & 9.3 & 11.3 \\
 & Bag-based & 7.7 & 5.7 \\
\midrule
 \multirow{2}{*}{\begin{tabular}[c]{@{}l@{}}Mean rank \\ (UAR)\end{tabular}} & Instance-based & 10.0 & 11.0 \\
 & Bag-based & 7.0 & 6.0 \\
\bottomrule
\label{Table:WR}
\end{tabular}}
\end{table}

The results for bag classification are reported in Fig. \ref{fig:WRLB} and \ref{fig:WRPB}. For an easier comparison between instance- and bag-based methods, mean ranks for all experiments are reported in Table \ref{Table:WR}. These results show that, \textbf{in general, bag-level methods outperform their instance-based counter-parts at higher WR} ($\geq 50\%$). At lower WR ($5\sim10\%$), the difference between both approaches is lower. However, in the Letters experiment, MI-SVM outperform all other methods by a significant margin, while in the HEPMASS experiment, EMD-SVM and NSK-SVM perform better. This suggests that \textbf{at lower WRs, there are other factors to consider when selecting a method}, such as the shape of the positive and negative distributions and the consistency of the WR across positive bags.

\subsection{Data Distribution: Non-Representative Negative Distribution}
\label{Section:Negative}

\begin{figure}
\centering
\includegraphics[width=0.75\textwidth]{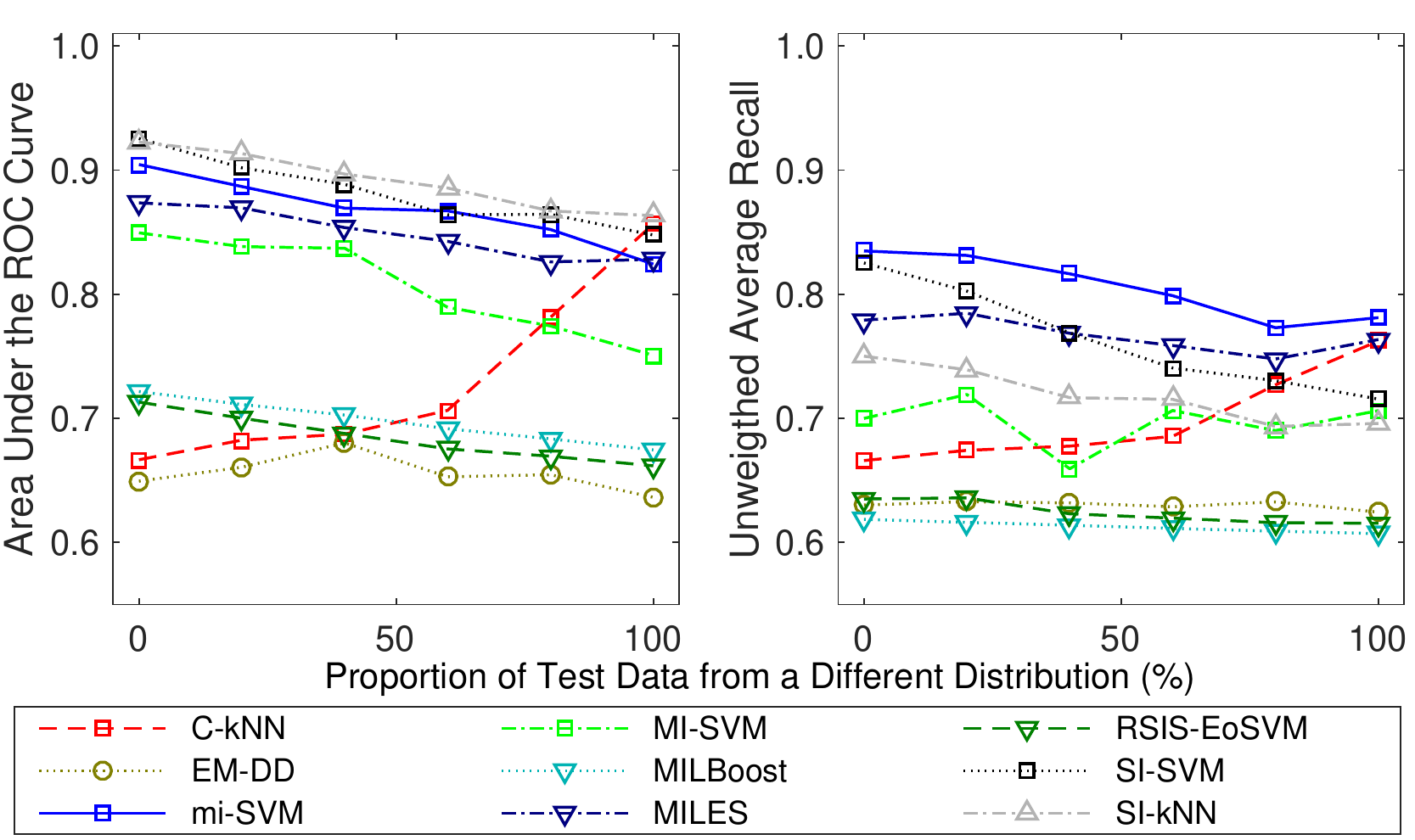}
\caption{Average performance of the MIL algorithms for instance classification on the Letters data as the test negative instance distribution increasingly differs from the training distribution.}
\label{fig:UNILI}       

\centering
\includegraphics[width=0.75\textwidth]{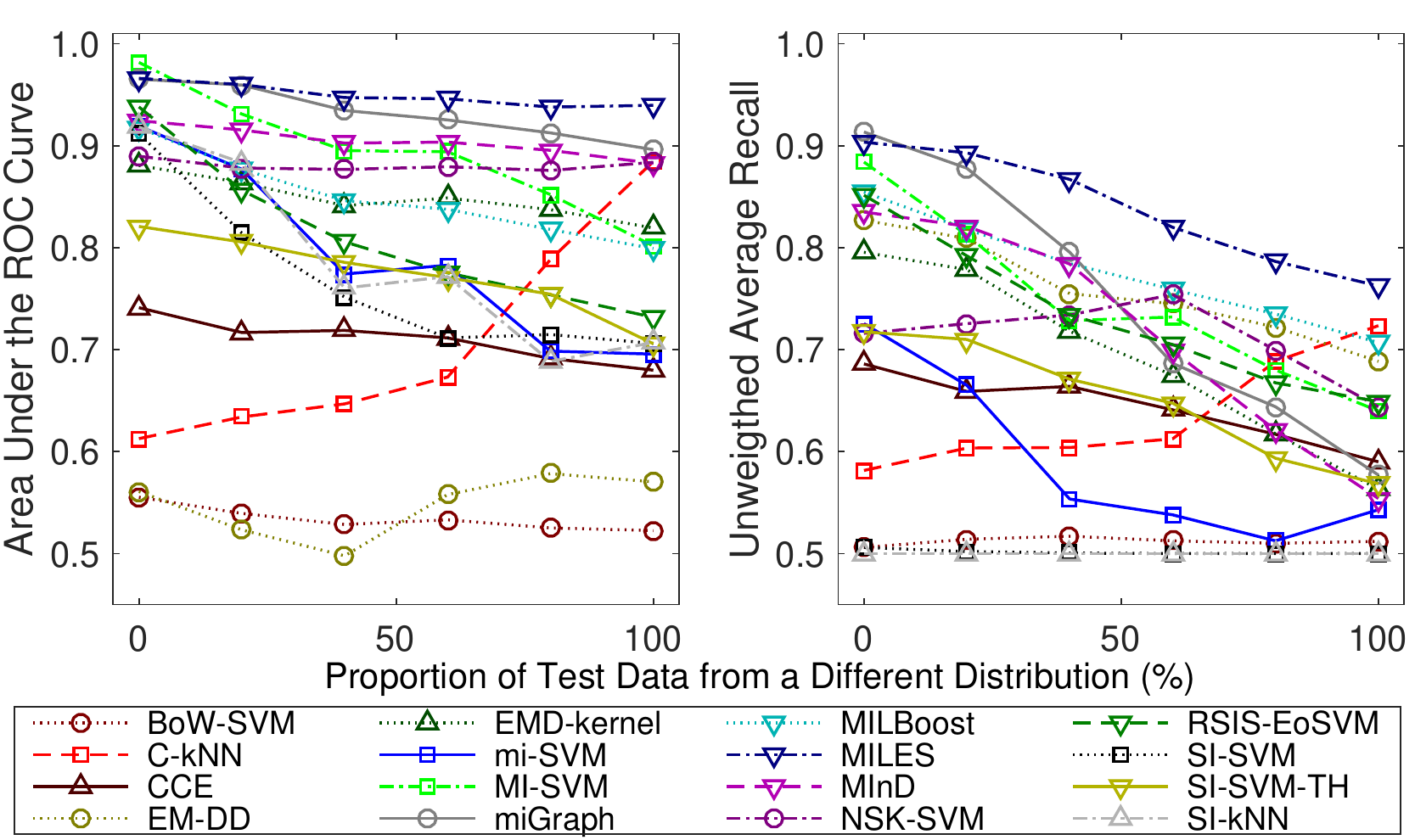}
\caption{Average performance of the MIL algorithms for bag classification on the Letters data as the test negative instance distribution increasingly differs from the training distribution.}
\label{fig:UNILB}       
\end{figure}


\begin{figure}
\centering
\includegraphics[width=0.75\textwidth]{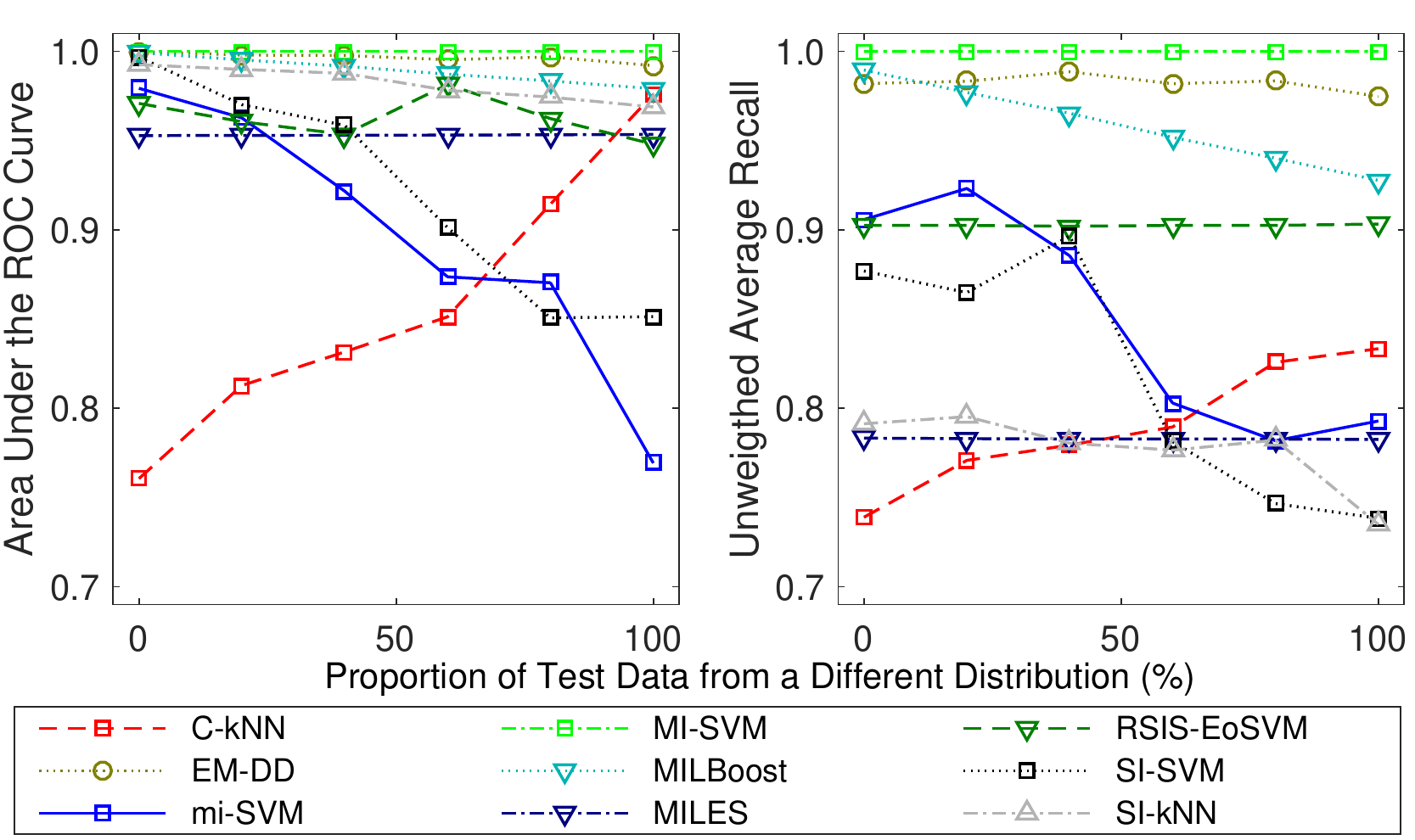}
\caption{Average performance of the MIL algorithms for instance classification on Gaussian toy data as the test negative instance distribution increasingly differs from the training distribution.}
\label{fig:UNITI}       

\centering
\includegraphics[width=0.75\textwidth]{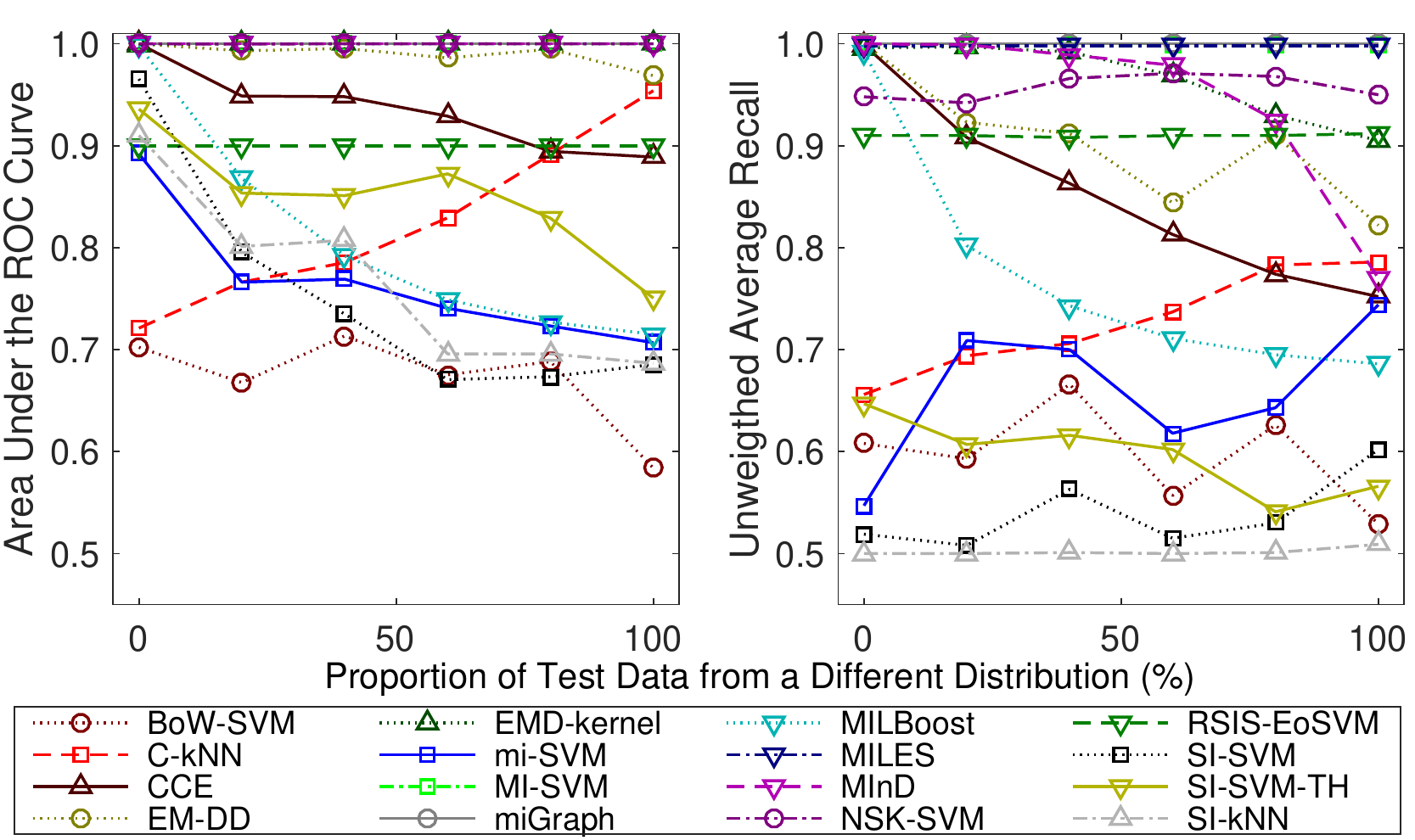}
\caption{Average performance of the MIL algorithms for bag classification on Gaussian toy data as the test negative instance distribution increasingly differs from the training distribution.}
\label{fig:UNITB}       
\end{figure}

In some applications, the negative instance distribution cannot be entirely represented by the training data set. The experiments in this section measure the ability of MIL algorithms to deal with a negative distribution different in test and training. Two data sets are used for these experiments: the Letters data set and the synthetic Gaussian toy data set created specially for this experiment. In each experiment, there are two different negative instance distributions. The first one is used to generate the training data. For the test data sets, at first, the negative instances are also sampled from this same distribution, but are gradually replaced by instances from the second distribution. The positive instances are sampled from the same distribution in both the training and test sets. For instance, using the Letters data set, this means that in the training data set the letter A, B and C are used as negative instances. Gradually, the instance from A, B and C are replaced by instance on the letter D, E and F.

The results of the experiments, illustrated in Fig. \ref{fig:UNILI}, \ref{fig:UNILB}, \ref{fig:UNITI} and \ref{fig:UNITB}, show that \textbf{most algorithms have decreasing performance when the test negative instances distribution differs from the training distribution}. However, C-kNN exhibits a contrasting behavior. More the test instances differ from test to training, the better are performances. This is because C-kNN uses the minimal Hausdorff distance as a similarity metric between bags. This is the distance between the two closest instances from each bag. If the negative instances come from the same distribution in all the bags, it is likely that the closest instance are both from the negative distribution, even if the bags are positive. If the bags have different labels, this leads to misclassification. If the negative test instances are different from those in the training set, the distance between two negative instances is likely to be greater than the distance between two positive instances, which are from the same distribution in both sets. Thus, positive bags are found to be closer to other positive bags leading to a higher accuracy.

The results for both data sets suggest that \textbf{bag-level methods are better for dealing with new negative distributions}. This may contribute to their success in computer vision applications. In Fig. \ref{fig:UNILB} the AUC for bag classification is stable for most method while their accuracy decreases. This suggest that the score functions learned by the algorithms are still suitable for the new distribution, but the thresholds should be adjusted. This observation motivates the use of adaptive methods in practice which would adjust the decision threshold as new data arrives.

\label{Section:UNI}

\section{Discussion}
\label{Section:Discussion}
The problem characteristics identified in this paper allow for a discussion on validation procedures of MIL algorithms. These suggestions are also based on the observations from the experiments in the previous section. Then, we identify interesting research avenues for MIL.

\subsection{Benchmarks Data Sets}
Several characteristics inherent to MIL problems were discussed in this paper. Experiments confirmed what has been observed by many researchers before: algorithms perform differently depending on the type of MIL problem, and several characteristics define a MIL problem. However, even to this day, many approaches are validated only with the Musk and Tiger/Elephant/Fox (TEF) data sets. There are several problems with these benchmark data sets. First, they pose only some of the challenges discussed earlier. For example, the WR of these data sets  is high. Since the instance labels are not supplied, the real WR is unknown. However, it has been estimated in some papers \cite{Gehler2007,Li2013,Li2010} which reported 82 to 100\% for Musk1, 23 to 90\% for Musk2 and  38 to 100\% for TEF. Moreover, in the Musk data sets, there are no explicit structure to be exploited. In the TEF data sets, the instances are represented by 230-dimensional feature vectors characterizing by color, texture and shape descriptors. No further details are given on these features, except that this representation is sub-optimal and should be further investigated \cite{Andrews02}. It is possible that the theoretical Bayesian error has already been reached for this feature representation and that better results are obtained on account of protocol related technicality, such as fold partitions. Also, since the annotations at instance level are not available, it is difficult to assess if the fox classifier really identifies foxes, or if it identifies background elements related to foxes such as forest segments. This would explain the high WR estimated in \cite{Li2013,Li2010,Gehler2007}. Since the state-of-the-art accuracy on this class is around 70\%, it is plausible that a large proportion of the animals in the negative class live in deserts or under the sea. For all these reasons, in our opinion, while the Musk and TEF data sets are representative of some problems, using more diverse benchmarks would provide a more meaningful comparison of MIL algorithms. 

Because of the aforementioned TEF shortcomings, researchers should use more appropriate benchmark data for computer vision tasks. For example, several methods have been compared on the SIVAL data set. It contains different objects captured in the same environments, and provides labels for instances. In each image the objects of interest are segmented into several parts. The algorithms ability to leverage co-occurrence can thus be measured, and since the objects are all captured in the same environments, the background instances do not interfere in the classification process. However, it would be more beneficial for the MIL community to use other existing strongly annotated computer vision data sets (e.g. Pascal VOC \cite{everingham2010pascal} or ImageNet \cite{Russakovsky2015}) as benchmarks. These types of data set provide bounding box or even pixel-level annotations that can be used to create instance labels in MIL problems. MIL algorithms could be compared to other types of techniques, which is almost never done in the MIL literature. Also, supplying the position of instances in images for these new computer vision MIL benchmarks would help to develop and compare methods that leverage spatial structure in bags. 

In application fields other than computer vision, there are relatively few publicly available real-world data sets. From these few data sets, to our knowledge, there is only one (Birds \cite{Briggs2012}) that supply instance labels and is non-artificial. This is understandable since MIL is often used to avoid the labor-intensive instance labeling process. Nevertheless, real-world MIL data needs to be created to measure the instance labeling capability of different MIL methods, as it is an increasingly important task. Also, to our knowledge, there is no publicly available benchmark data set for MIL regression, which would surely stimulate research on this task.

Finally, several methods are validated using semi-artificial data sets. These data sets are useful to isolate one parameter of MIL problems, but are generally not representative of real-world data. In these data sets, instances are usually i.i.d. which almost never happens in real problems. Authors should justify the use of this type of data, clearly mention what assumptions are made and how the data sets are different from real data. As a start, Table \ref{Table:DataSets} compiles the characteristics which are believed to be associated with some of the most widely used benchmark data sets, based on parameter estimation and data descriptions found in literature. These are believed to be true but would benefit from rigorous investigation in the future.

In short, whenever only the Musk and the TEF data sets are used to validate a new method, it is difficult to predict how the methods will perform in different MIL problems. Moreover, because researchers are encouraged to evaluate their methods on these data sets, promising models may be dismissed too early because they do not outperform the best performing methods optimized on these benchmark data sets. We argue that a better understanding of the characteristics of the MIL data sets should be promoted, and that the community should use other data sets to compare MIL algorithms in regard of the challenges and properties of MIL problems. 

\begin{table} \centering
\caption{Table compiling the characteristics of MIL benchmark data sets based on statement in the literature.}

\scalebox{0.9}{
\begin{tabular}{l|c|c|c|c|c|c|c|c|c|}
\textbf{Benchmark MIL Data Sets} & \rot{Instance labels} & \rot{Low witness rate} & \rot{Intra-bag similarities} & \rot{Instance co-occurence} 
& \rot{Structure in bags} & \rot{\shortstack[l]{Multimodal positive\\distribution}} & \rot{\shortstack[l]{Non-representative\\negative distribution}} 
& \rot{Label noise} & \rot{Semi-Artificial} \\
\cmidrule{1-10}

Musk \cite{Dietterich1997}
& 	&	&\OK &   &   & \OK & \OK &   & \\
Tiger, Fox, Elephant \cite{Andrews02} 
& & &\OK & \OK & & \OK & \OK & & \\
SIVAL \cite{Settles2008}
& \OK & & & & &\OK & & &\\
Birds \cite{Briggs2012}
& \OK & \OK & & \OK & & & & & \\
Newsgroups \cite{Settles2008}
& \OK & \OK & & & & \OK & & &\OK\\
Corel \cite{Chen2004DDSVM}
& & &\OK & \OK & & \OK & \OK & & \\
Messidor Diabetic Retinopathy \cite{Kandemir2015} 
& &\OK & & & & \OK & & &\\
UCSB Breast \cite{Kandemir2014empowering}
& &\OK & & & & \OK & & &\\
Biocreative \cite{Ray2005}
& & & \OK & \OK & & \OK &  & &\\
\cmidrule[1pt]{1-10}
\end{tabular}}
\label{Table:DataSets}
\end{table}

\subsection{Accuracy vs. AUC}
While benchmark data is of paramount importance, the proper selection of performance metrics is equally important to avoid hasty conclusions. In all experiments, some algorithms have obtained contrasting performance when comparing AUC to accuracy and UAR. This has also been observed in other experiments \cite{Carbonneau2016}. This is an important factor that must be taken into consideration when comparing MIL algorithms. 

Some algorithms (e.g. mi-SVM, SI-kNN, SI-SVM, miGraph, MILES) obtain high AUC that does not translate into high accuracy. There may be many reasons for this. Some algorithms optimize the decision thresholds based on bag accuracy, while others infer individual instance labels. In the first case, the algorithm is more prone to FN, while the latter is more prone to FP because of the asymmetric misclassification costs discussed in Section \ref{Section:IC}. Figure \ref{fig:UNILB} and Figure \ref{fig:UNITB} in Section \ref{Section:Negative} clearly illustrate this. As the negative distribution changes, the AUC remains stable for many algorithm, while accuracy decreases (e.g. miGraph, MILES, BoW-SVM). This means that the score function was still suitable for classification, but the decision threshold was no longer optimal. Considering the right end of the AUC curves in Figure \ref{fig:UNILB}, where negative instances are completely sampled from a new distribution, one could conclude that miGraph performs better than RSIS-EoSVM. However, when comparing with UAR, the inverse can be concluded. One could argue the AUC is a sufficient performance metric assuming that the decision threshold is optimized on a validation set, however, in many problems, the amount of available data is too limited for this assumption to hold. Also in the case of instance classification, instance labels are unknown, therefore, it is not possible to perform such optimization. 

In our opinion, the algorithms ability to accurately set this threshold is an important characteristic that should be measured, as well as the ability to learn a suitable score function. Therefore, accuracy should always be reported alongside AUC.

\subsection{Future Direction}
Based on the literature review of this survey, we identify several MIL topics that are interesting avenues for future research. 

First, tasks like regression and clustering are not extensively studied when compared to classification. This might be because there are less applications for these tasks, and because there are no publicly available data. A good place to start exploration on MIL regression could be in affective computing applications, where the objective is to quantify abstract concepts, such as emotions and personalities. In these applications, real-valued labels express the appreciation of human judges for speech or video sequences (bags). The sequences are represented by an ensemble of observations (instances), and it is unclear which observation contributed to the appreciation level. In this light, these problems perfectly fit in the MIL framework. Better regression algorithms would also be useful in CAD to assess the progression stage of a pathology instead of only classifying subjects as diseased or healthy. 

Also, it is only fairly recent that the difference between instance and bag classification is throughly investigated. It is demonstrated in \cite{Doran2014,Vanwinckelen2015}, in Section \ref{Section:IvsB} and our experiments that these tasks are different. It is showed in this paper and \cite{Carbonneau2016IPTA} that many instance-space methods proposed for bag classification are sub-optimal for instance classification. There is a need for MIL algorithms primarily addressing instance classification, instead of performing it as a side feature. Based on the results Section \ref{Section:ExpIC} approaches discarding or only minimally using the bag arrangement information appears to be better suited for this task. We believe that this bag arrangement could be better leveraged than how it is done by existing methods, which often seek to maximize bag-level accuracy. To further stimulate research on this topic, more instance-annotated MIL data sets are needed.     

While tasks outside bag classification would benefit from more exploration, there are also problem characteristics that necessitate the attention of the MIL community. For instance, intra-bag similarities have never been identified as a challenge, and thus, directly addressed. It could be beneficial to perform some sort of normalization or calibration in each bag to remove what is common to each instance and specific to the bag. In computer vision, this is usually done in a preliminary normalizing step. However, in other tasks such as molecule classification, this type of procedure could be helpful. For example, in the Musk data, the instances in the bag are conformations of the same molecule. Discarding the information related the ``base'' shape of the molecule could help to infer what more subtle particularity of the configurations is responsible for the effect when comparing to other molecules.   

There are only a few methods that leverage the structure in bags. This is an important topic that has been addressed in some BoW methods, but was never thoroughly looked upon in other types of MIL methods, except for some methods using graphs \cite{Zhou2009migraph,Yan2016,Zhang2011MILSD,Mcgovern2003}. Some of these methods represent similarities between instances or represent whole bag as graph. Methods that create an intermediate graph representation in which some instances are grouped in sub-graphs could be an interesting way to leverage the inner structure of bags. In that case, the witness would be an ordered arrangement of instances. With this type of representation, complex objects could be identified more reliably in complex environments. 

In many problems, the numbers of negative and positive instances are severely imbalanced, and yet, the existing learning methods for imbalanced data set have not studied extensively in MIL. There exist many methods to deal with imbalanced data \cite{Branco2016}. There are external methods like SMOTE \cite{Chawla2002} and RUSBoost \cite{Seiffert2010} that necessitate accurate labels to perform over or under sampling. To be adapted to MIL these methods could use some kind of probabilistic label function. Internal methods \cite{Imam2006,Veropoulos1999} adjust the misclassification cost independently for each class. These schemes could be used in algorithms such as mi-SVM which require the training of an SVM with high class imbalance when the WR is low. Class imbalance has also been identified in \cite{Herrera2016MILBook} as an important topic for future research. 

There are other MIL challenges that were not studied in this paper due to space constraints. For one, the computational complexity of the algorithms is important since MIL is often used to leverage large quantities of data. This is generally not a concern for bag-level methods. However, instance-level methods rapidly become difficult to use with large data sets. The elaboration of methods focused on computational efficiency would facilitate the use of MIL in large-scale applications. 

When working with MIL, one must deal with uncertainty. It would be beneficial in many applications to use active learning to train better classifiers by querying humans about most uncertain parts of the feature space. For example, in CAD, after preliminary image classification, the algorithm would determine which are the most critical instances and prompt the clinician to provide a label. These critical instances would be the most ambiguous or the ones that would most help the classifier. This would necessitate research to assert degrees of confidence in parts of feature space. Alternatively, the algorithm should be able to evaluate the information gain that each instance label would provide. As a related topic, new methods should be proposed to incorporate knowledge from external and reliable sources. Intuitively, the information obtained with strong labels should have more importance in the MIL algorithm's learning and decision process than instance with weak labels.   

Except for a few papers, MIL methods always focus on classification/regression, and features are considered as immutable parameters of the problem. Recently, methods for representation learning \cite{Bengio2013} have gained in popularity because they usually yield a high level of accuracy. Some of these methods learn features in a supervised manner to obtain a more discriminative representation \cite{Mairal2008}, or, in deep learning, a supervised training phase is often used to fine tune the features learned in an unsupervised manner \cite{Larochelle2009}. This cannot be done directly in MIL because of the uncertainty on the labels. The adaptation of discriminative feature learning methods would be beneficial to MIL. Also, it has be shown that mid-level representation help to bridge the semantic gap between low-level features and concepts \cite{Hauptmann2007, Li2010objectBank,Sadanand2012}. These methods obtain a mid-level representation using supervised learning on images or videos annotated with bounding boxes. Learning techniques for these mid-level representations should also be proposed for MIL. This is an area where multiple instance clustering would be useful. There are already a few papers on this promising subject \cite{Zhu2013,Zhu2015objDisc}. However, there are still a lot of open questions and limitations to overcome, such as dealing with multiple objects in a single image or the dependency to a saliency detector.

In some applications, like emotion or complex event recognition from videos, objects are represented using different modalities. For example, the voice and facial expression of a subject can be used to analyze its behavior or emotional state \cite{Ringeval2013}. Alternatively, events in videos can be represented, among others, by frame, texture and motion descriptors \cite{Merler2012,Tang2013}. In both cases, a video sequence is represented by a collection of feature vectors, which corresponds to a bag in MIL. The difference with existing MIL problems is that these instances belong to a different feature spaces. This is an interesting problem that has yet to be addressed by the MIL community. This will be useful in rising research areas, such as multimedia analysis or problems related to the Internet-of-things, which necessitate the fusion of diverse sources of information. By their nature these applications imply large quantity of data, and thus MIL would be a perfect tool to leverage all this information and reduce the burden of annotation. Several fusion strategies should be explored. Instance could be mapped to the same semantic space to be compared directly, graph model could be used to aggregate several heterogeneous descriptors or instances could be combined in pairs to create new spaces for comparison similarly to \cite{Daume2009}.

\section{Conclusion}
\label{Section:Conclusion}
In this paper, the characteristics and challenges of MIL problems were surveyed with applications in mind. We identified four types of characteristics which define MIL problems and dictate the behavior of MIL algorithms on data sets. It is an important topic in MIL because a better knowledge of these MIL characteristics helps interpreting experiments results and may lead to the proposal of improved methods in the future. 

We conducted experiments using 16 methods which represent a broad spectrum of approaches. The experiments showed that these characteristics have an important impact on performance. It was also shown that each method behaves differently given the problem characteristics. Therefore, careful characterization of problems should not be neglected when experimenting and proposing new methods. More specific conclusions have also been drawn from experiments:   

\begin{itemize}
\item For instance classification tasks, when the WR is relatively high, there is no need for MIL algorithms. The problem can be cast as a regular supervised problem with one-sided noise.
\item For instance classification tasks, the best approaches do not use use bag information (or only very lightly). Also, methods optimized using bag classification accuracy as an objective have a higher false negative rate (as the WR increases), which limits their performance for this task.
\item Bag-level methods and methods assuming instances inherit their bag label yield better classification performance especially when the WR is high. 
\item Bag-space methods are more robust than instance-space methods in problems where the negative distribution cannot be completely represented by the training data. This was particularly true when using the minimal Hausdorff distance. 
\item Measuring performance only in terms of AUC is misleading. Some algorithms learn an accurate score function, but fail to optimize the decision threshold used to obtain hard labels, and thus, yield low accuracy.

\end{itemize}

After observing how problem characteristics impact MIL algorithms, we discussed the necessity of using more benchmark data sets than the Musks and Tiger, Elephant and Fox data sets to compare proposed MIL algorithms. It became evident that appropriate benchmark data sets should be selected based on the characteristics of the problem to be solved. We then identified promising research avenues to explore in MIL. For example, we found that only few papers address MIL regression and clustering, which is useful in emerging applications such as affective computing. Also, more methods leveraging structure among instances should be proposed. These methods are in high demand in the era of the Internet of things, where large quantities of time series data are generated. Finally, methods dealing efficiently with large amount of data, multiple modalities and class imbalance require further investigation.






\bibliographystyle{IEEETran}
\bibliography{MIL1}

\end{document}